\documentclass[journal]{IEEEtran}

\usepackage{microtype}
\usepackage{graphicx}
\usepackage{subfigure}
\usepackage{booktabs} 
\usepackage{hyperref}

\usepackage{times}
\usepackage{soul}
\usepackage{url}
\usepackage[utf8]{inputenc}
\usepackage[small]{caption}
\usepackage{graphicx}
\usepackage{wrapfig}
\usepackage{amsmath}
\usepackage{amsthm}
\usepackage{booktabs}
\usepackage{algorithm}
\usepackage{algorithmic}
\usepackage[american]{babel}
\usepackage{microtype}

\usepackage{amsfonts}
\usepackage{amsmath,bm,paralist}
\usepackage{epstopdf}
\usepackage{subfigure}
\usepackage[T1]{fontenc}
\usepackage{makecell}
\usepackage{comment}
\usepackage{enumitem}
\allowdisplaybreaks

\def \I {\mathcal{I}}
\def \O {\mathcal{O}}
\def \x {\mathbf{x}}

\def \w {\mathbf{w}}
\def \R {\mathbb{R}}

\def \P {\mathbf{P}}

\DeclareMathOperator*{\rk}{rank}

\DeclareMathOperator*{\E}{E}

\DeclareMathOperator*{\argmin}{argmin}

\newtheorem{thm}{Theorem}
\newtheorem{lem}{Lemma}

\theoremstyle{definition}
\newtheorem{remark}{Remark}

\begin{document}

\title{Prediction with Unpredictable Feature Evolution}

\author{
Bo-Jian Hou, Lijun Zhang,~\IEEEmembership{Member,~IEEE} and~Zhi-Hua Zhou,~\IEEEmembership{Fellow,~IEEE}
\thanks{All authors are with the National Key Laboratory for Novel Software
Technology, Nanjing University, Nanjing 210023, China. Z.-H. Zhou is the corresponding author. E-mail: \{houbj,zhanglj,zhouzh\}@lamda.nju.edu.cn.}
\thanks{This work was supported by NSFC (61921006) and Collaborative Innovation Center of Novel Software Technology and Industrialization.}
}

\markboth{IEEE Transactions on Neural Networks and Learning Systems}%
{Shell \MakeLowercase{\textit{et al.}}: Bare Demo of IEEEtran.cls for IEEE Journals}

\maketitle

\begin{abstract}

Learning with feature evolution studies the scenario where the features of the data streams can evolve, i.e., old features vanish and new features emerge. Its goal is to keep the model always performing well even when the features happen to evolve. To tackle this problem, canonical methods assume that the old features will vanish simultaneously and the new features themselves will emerge simultaneously as well. They also assume there is an overlapping period where old and new features both exist when the feature space starts to change. However, in reality, the feature evolution could be unpredictable, which means the features can vanish or emerge arbitrarily, causing the overlapping period incomplete. In this paper, we propose a novel paradigm: \emph{Prediction with Unpredictable Feature Evolution} (PUFE) where the feature evolution is unpredictable. To address this problem, we fill the incomplete overlapping period and formulate it as a new matrix completion problem. We give a theoretical bound on the least number of observed entries to make the overlapping period intact. With this intact overlapping period, we leverage an ensemble method to take the advantage of both the old and new feature spaces without manually deciding which base models should be incorporated. Theoretical and experimental results validate that our method can always follow the best base models and thus realize the goal of learning with feature evolution.

\end{abstract}

    
\IEEEpeerreviewmaketitle

\section{Introduction}
\label{section:introduction}
In the big data era, data often come in a streaming way since the data often have big volume and high velocity. Learning with data streams has been studied extensively~\cite{DBLP:conf/kdd/DomingosH00,DBLP:conf/edbt/SeidlAKKH09,DBLP:conf/ijcnn/LeiteCG09, DBLP:conf/icml/TsangKK07}, where the typical methods in the literature assume a fixed set of features. However, in a practical scenario, the features of the data streams often evolve, i.e., old features vanish and new features emerge. For example, in ecosystem protection, people deploy sensors in the ecosystem to collect data, where each sensor corresponds to a feature. Due to deterioration or unexpected damage, after some time, many sensors will be out of use, and new sensors will be deployed. This scenario also occurs in object recognition or indoor surveillance~\cite{DBLP:conf/infocom/WangX0XL16}. 

\begin{figure}[!t]
    \centering
    \includegraphics[width=1\linewidth]{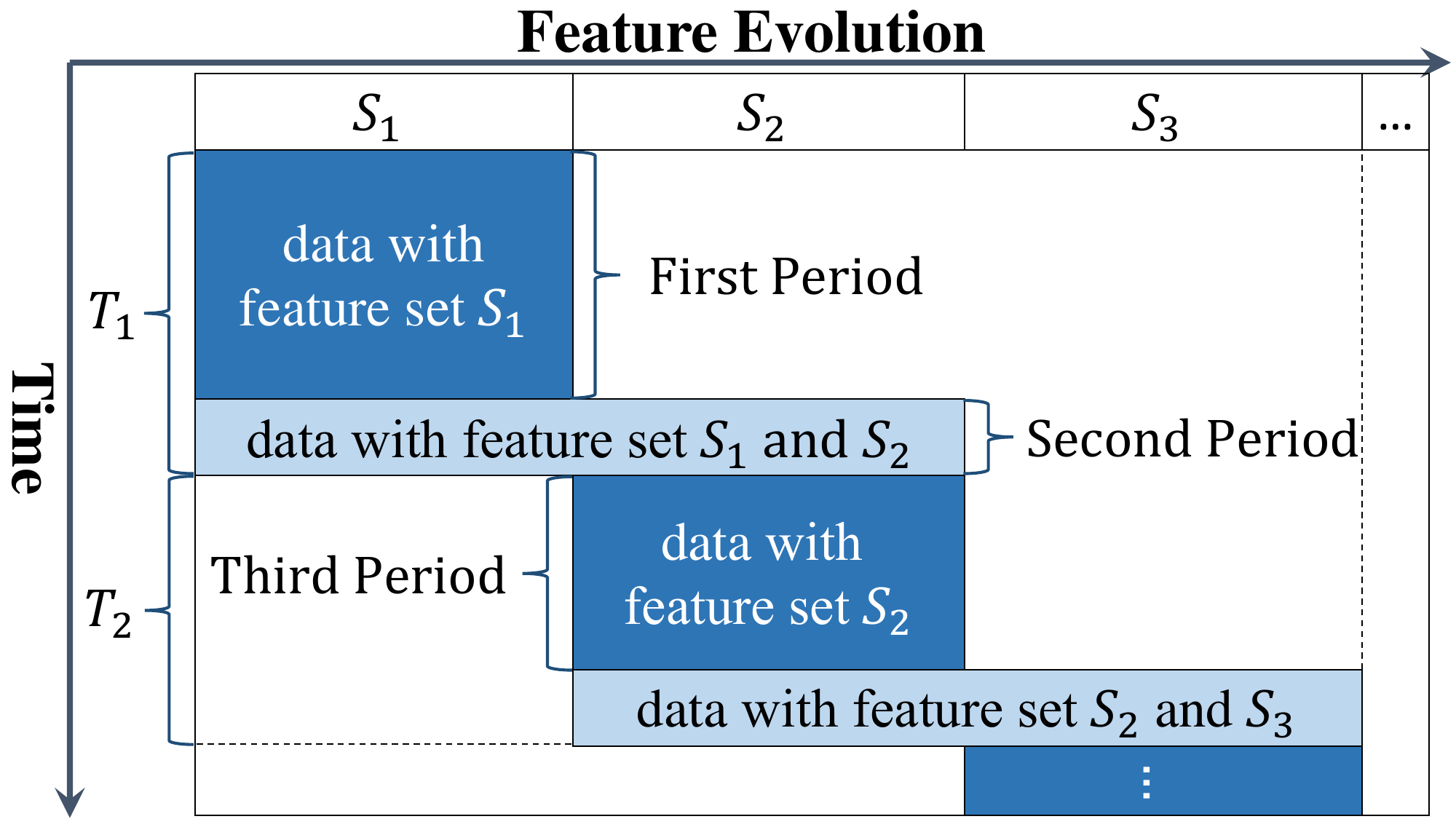}
    \caption{Illustration of how data streams come in FESL. $S_1$, $S_2$ and $S_3$ are different feature sets. $T_1$ is the period where $S_1$ is valid and $T_2$ is the period where only $S_2$ is valid. At the end of $T_1$, namely the second period, samples are described by both $S_1$ and $S_2$. The second period is also called the {\it overlapping period}.}
    \label{Illustration-general}
    \vspace{-0.5cm}
\end{figure}

When the features start to evolve, since there are only limited samples described by these new features, it is not sufficient to train a strong model based on these samples. Besides, the samples described by the old features are ignored, which is a big waste of the data collection effort. To tackle this problem, FESL (Feature Evolvable Streaming Learning)~\cite{DBLP:conf/nips/Hou0Z17} assumes that features do not change in an arbitrary way and instead
there are some overlapping periods in which both old and
new features are available. Figure~\ref{Illustration-general} illustrates how the data streams come in FESL. 


\begin{figure}[!t]
    \centering
    \includegraphics[width=1\linewidth]{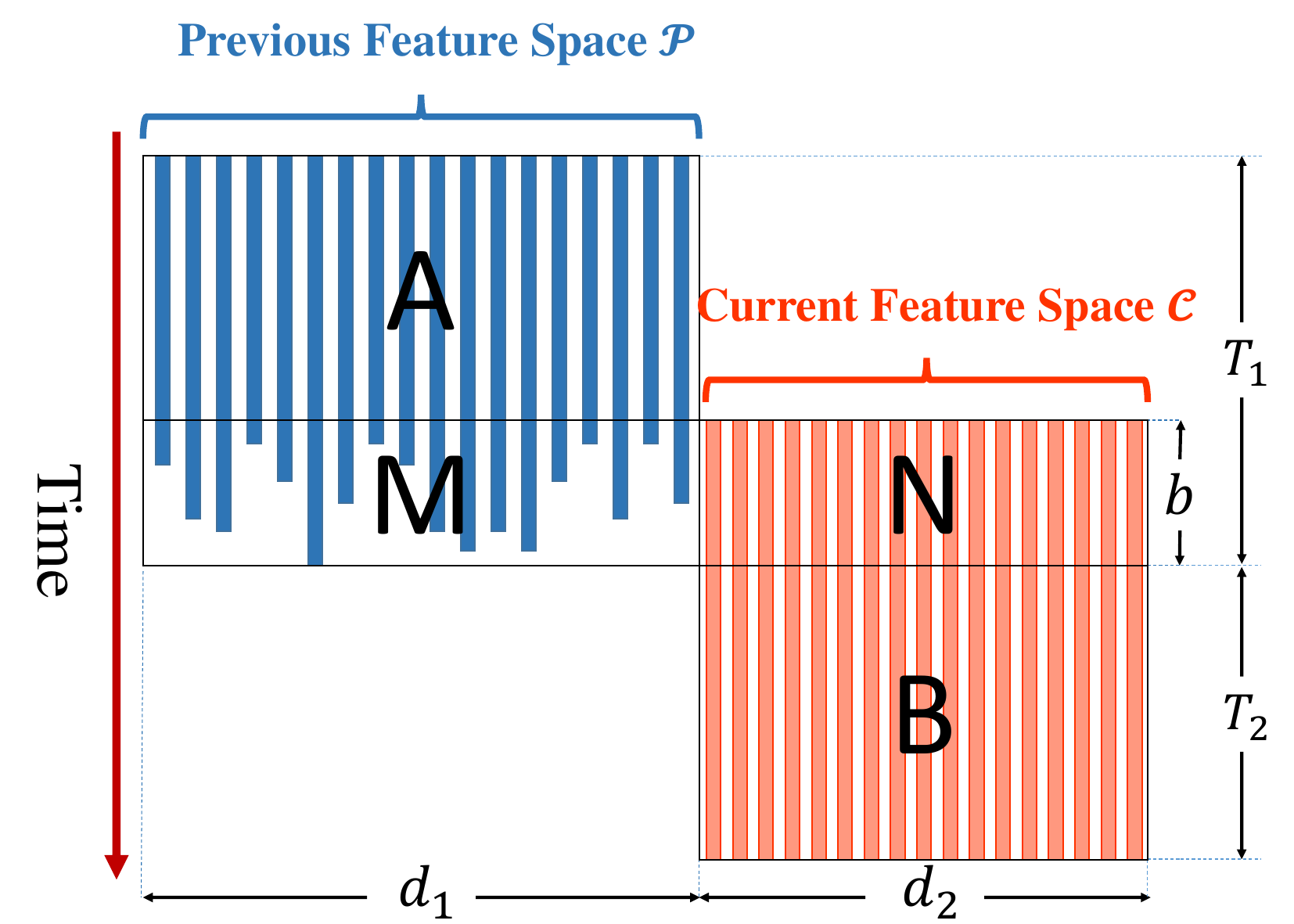}
    \caption{Illustration of PUFE with two feature spaces. The previous feature space and the current feature space are denoted by $\mathcal{P}$ and $\mathcal{C}$ respectively. $T_1$ is the period where $\mathcal{P}$ is valid, $b$ is the overlapping period where both feature spaces are available, $T_2$ is the period where only $\mathcal{C}$ is valid. $d_1$ and $d_2$ are the dimensions of $\mathcal{P}$ and $\mathcal{C}$ respectively. $A$, $M$, $N$, $B$ are the matrices formed by corresponding samples, in which $M$ is incomplete due to the unpredictable feature evolution.} 
    \label{Illustration}
\end{figure}

From Figure~\ref{Illustration-general} we can see that the features in the same feature set vanish at the same time. However, in reality the FESL's assumption may be too strong where the feature evolution can be unpredictable. Back to the ecosystem protection example, due to the different situations of sensors, such as the difference on positions, temperatures, magnitudes of signal, etc., the sensors' expiring time would be different. Therefore, the features corresponding to the sensors with short lifespans will vanish earlier than the others which is illustrated in Figure~\ref{Illustration}. On the other hand, generally it is reasonable to assume that new sensors are deployed simultaneously since it is much more efficient and can save workload than employing new sensors one by one. Thus the new features will appear simultaneously.



In this paper, we propose a novel paradigm: \emph{Prediction with Unpredictable Feature Evolution} (PUFE) where old features vanish unpredictably and new features emerge simultaneously. We define the ``feature space" in our paper by the feature set. Figure~\ref{Illustration-general} shows that the three periods form a cycle and each cycle merely includes two feature spaces. Thus, we only need to focus on one cycle and it is easy to extend to the case with multiple cycles. Figure~\ref{Illustration} gives the illustration of how data streams come in PUFE. We call the two feature spaces ``previous'' and ``current'' feature space with notations $\mathcal{P}$ and $\mathcal{C}$ respectively. Each column represents a feature. According to Figure~\ref{Illustration}, the process of PUFE can be summarized as follows:
\begin{itemize}
    \item For $t=1,\ldots,T_1-b$, in each round, the learner observes a vector $\x_t^P\in\mathbb{R}^{d_1}$ sampled from $\mathcal{P}$ where $d_1$ is the number of features of $\mathcal{P}$, $T_1$ is the number of total rounds in $\mathcal{P}$ and $b$ is the number of the rounds in overlapping period. All observed data form matrix $A$.
    \item For $t=T_1-b+1,\ldots,T_1$, in each round, the learner observes a portion of vector $\x_t^P\in\mathbb{R}^{d_1}$ from $\mathcal{P}$ that finally form the incomplete matrix $M$ and the intact vector $\x_t^C\in\mathbb{R}^{d_2}$ from $\mathcal{C}$ that finally form the intact matrix $N$. $d_2$ is the number of features of $\mathcal{C}$. 
    \item For $t=T_1+1,\ldots,T_1+T_2$, in each round, the learner observes vector $\x_t^C\in\mathbb{R}^{d_2}$ sampled from $\mathcal{C}$ where $T_2$ is the number of rounds in $\mathcal{C}$. All observed data form matrix $B$. Note that $b$ is small, so we can omit the data from $\mathcal{C}$ on rounds $T_1-b+1,\ldots,T_1$ since they have minor effect on training the model in $\mathcal{C}$.
\end{itemize}

To address the problem of unpredictable missing features, we impute the missing value of $M$ by reducing the original problem into a matrix completion problem in which the samples are observed without replacement. We prove a theoretical bound on the least number of observed entries that is far more less than that of the conventional methods with the help of matrix $A$. Then we use the intact overlapping period (reconstructed $M$ together with $N$) to learn the mapping from $\mathcal{C}$ to $\mathcal{P}$ so as to recover the data from $\mathcal{P}$ when there are only data from $\mathcal{C}$. With predictions of the old model on the recovered data, we propose a new ensemble method to make our model always comparable with the best base models that are only trained in the single feature space and thus solve the issue of learning with feature evolution: always keep the model performing well even when the feature evolving happens. Furthermore, this ensemble method do not need to manually decide which base models should be incorporated when newer feature spaces come. In summary, our major contributions are as follows:

\begin{enumerate}
\item we propose a more practical setting PUFE, where the old features will vanish unpredictably as feature evolves; 
\item we formulate the unpredictable evolution as a new matrix completion problem and propose an effective method with much smaller observed entries than conventional ones;
\item we propose to leverage the assistance from the previous feature space and theoretically guarantee that our model is always comparable to the best baseline and can adaptively tackle the situation when newer feature space appears;
\item the experimental results show that our model is comparable to the best baseline and surprisingly better than them in most cases, which validate the effectiveness of PUFE;
\item our algorithms are in a one-pass manner without saving any data, which is very valuable in learning with data streams since it is infeasible to keep the whole data due to the streaming nature.
\end{enumerate}
In the following, Section~\ref{section:setting} provides the framework of PUFE. The proposed approach with corresponding theoretical guarantees are presented in Section~\ref{section:proposed approach}. Section~\ref{section:experiments} reports the experimental results. Section~\ref{section:Related Work} introduces related works. Finally, Section~\ref{section:conclusion} concludes our paper.

\section{Framework}
\label{section:setting}

We focus on both classification and regression tasks. In each round, the predictor receives an instance and is required to do prediction. After the prediction has been made, the true label is revealed and the predictor will suffer a $loss$ which exhibits the discrepancy between the prediction and the true label.

Our goal is to always keep the model performing well even when there are only few data when new feature space emerges. It is worth noting that this goal should be implemented without storing the history or in a one-pass manner due to the large volume and high velocity of data. Specifically, we want to make our model obtain good performance in the current feature space during period $T_2$ shown in Figure~\ref{Illustration}, no matter at the beginning or at any other time step. The basic idea is to establish relationship between the previous and current feature space by an overlapping period where both previous and current features exist. Then the well-learned model in the previous feature space can be utilized to assist the performance in the current feature space. Nevertheless, in our setting, we do not have an intact overlapping period. Thus we need to study whether we can rebuild it and this may need matrix completion techniques~\cite{DBLP:journals/jmlr/Recht11, DBLP:journals/cacm/CandesR12,JMLR:2019:Zhang}. Since time is seasonal, it is reasonable to assume that two instances on the same periodic point are linearly related. Thus, we have a chance to rebuild the overlapping period with the help of observed intact instances from the previous feature space, i.e., the matrix $A$ shown in Figure~\ref{Illustration}.

So far, the framework of our approach has been clear. Concretely, we have mainly four steps. The first step is to learn a good model in the previous feature space as a prepared backup. Then in order to build the relationship between the previous and the current feature space, we fill part of the overlapping period, that is the matrix $M$ shown in Figure~\ref{Illustration}, where the features start to vanish. In the third step, we learn a mapping between $M$ and $N$. Finally, we make predictions in the current feature space, which will be boosted by the well-learned model from the previous feature space by utilizing its prediction on the data recovered by the mapping. The framework of our method is summarized in Algorithm~\ref{framework}. It is worthy to emphasize that all the processes can be operated sequentially which means we do not need to store any data, which is valuable in learning with data streams.

Although it seems that there is only a small difference between FESL and PUFE, PUFE is indeed more practical than FESL and has more application value. Besides, the challenge brought by this difference is not simple for which we make several efforts and contributions. Concretely, completing matrix $M$ in step two is not trivial since the missing of items is not uniformly random but with certain rule. This will be discussed in Section \ref{section:complete}. We give a theoretical guarantee on the number of observed entries, which is much smaller than the conventional one. In addition, learning a good model through utilizing the assistance from the previous feature space in step four is also not easy since at the beginning, we should follow the good model, say $h$ learned from the previous feature space. However, there might be errors when doing recovering. Then after a period of time, this model $h$ would be worse and worse since more and more recovered errors accumulate. Thus we have to discard $h$ and follow new good base model adaptively. We provide a tighter bound than FESL, and besides that our model can be extended to tackle the situation adaptively when newer feature space appears. In other words, we do not need to decide manually which base model should be incorporated in and which base model should be discarded while FESL has to decide it manually. This will be discussed in Remark~\ref{remark:prediction}.

\begin{algorithm}[t]
\caption{Framework of PUFE}
\label{framework}
\begin{algorithmic}[1]
\STATE Learn a model sequentially in the previous feature space with \textbf{Algorithm~\ref{alg:Initialize}}.
\STATE Complete matrix $M$ sequentially shown in Figure~\ref{Illustration} with \textbf{Algorithm~\ref{alg:complete}}.
\STATE Learn a mapping sequentially from $N$ to $M$ using~(\ref{equation:linear least square}).
\STATE Make predictions sequentially in the current feature space with \textbf{Algorithm~\ref{alg:prediction}}.

\end{algorithmic}
\end{algorithm}

\section{The Proposed Approach: PUFE}
\label{section:proposed approach}
According to the framework presented in Algorithm~\ref{framework}, in this section, we use four subsections to present the detailed implementation of our proposed approach.
\subsection{Learn a Model from $A$ in Previous Feature Space}
\label{section:A}

We use $\|\x\|$ to denote the $\ell_2$-norm of a vector $\x$. The inner product is denoted by $\langle\cdot,\cdot\rangle$. Let $\O_P\subseteq \mathbb{R}^{d_1}$ be the set of linear models in the previous feature space that we are interested in. We define the projection $\Pi_{\O_P}(\textbf{b})=\argmin_{\textbf{a}\in\O_P}\|\textbf{a}-\textbf{b}\|$. We restrict our prediction function at $t$th round to be linear which takes the form $\langle\w_{P,t},\x_t^P\rangle$ where $\w_{P,t}\in\mathcal{O}_P$. The loss function $\ell(\w^\top\x,y)$ is convex in its first argument. In implementing algorithms, we use {\it logistic loss} for classification tasks, namely,
\begin{equation}
\label{equation:logistic loss}
\ell(\w^\top\x,y)=\ln(1+\exp(-y(\w^\top\x))),
\end{equation}
while in regression tasks, we use \emph{square loss}, namely,
\begin{equation}
\label{equation:square loss}
\ell(\w^\top\x,y)=(y-\w^\top\x)^2.
\end{equation}

We follow FESL~\cite{DBLP:conf/nips/Hou0Z17} to learn an online linear model from the previous feature space, i.e., matrix $A$ shown in Figure~\ref{Illustration} sequentially by online gradient descent~\cite{DBLP:conf/icml/Zinkevich03}. The model $\w_{P,t}$ is updated during rounds $1,\ldots,T_1-b$ according to: 
\begin{equation}
\label{equation:update model1}
\w_{P,t+1}=\Pi_{\O_P}\left(\w_{P,t}-\tau_t\nabla\ell(\w_{P,t}^\top\x_t^{P},y_t)\right),
\end{equation}
where $\tau_t$ is a varied step size.

The process of learning a model from $A$ during rounds $1,\ldots,T_1-b$ are concluded in Algorithm~\ref{alg:Initialize}. Specifically, we first initialize our linear model $\w_{P,0}\in\mathcal{O}_P$ randomly. Then in each round, the predictor receives an instance $\x_t^P\in\mathbb{R}^{d_1}$ from $\mathcal{P}$ where $d_1$ is the number of the dimension of $\mathcal{P}$. The predictor makes predictions on this instance by $p_t=\langle\w_{P,t},\x_t^P\rangle\in\mathbb{R}.$ After the prediction has been made, the true label $y_t\in\mathbb{R}$ is revealed and the predictor will suffer a loss $\ell(p_t,y)$ according to (\ref{equation:logistic loss}) or (\ref{equation:square loss}). Finally, based on this loss, the predictor will update itself using (\ref{equation:update model1}). We set the varied step size $\tau=1/\sqrt{t}$ which can derive a good theoretical bound in Theorem~\ref{thm:prediction}.

\begin{algorithm}[!t]
	\centering
	\caption{Learn a Model from $A$}
	\label{alg:Initialize}
	\begin{algorithmic}[1]
		\STATE Initialize $\w_{P,0}\in\O_P$ randomly.
		\FOR{$t=1,2,\ldots,T_1-b$}
	    \STATE Receive $\x_t^P\in\mathbb{R}^{d_1}$ and predict $p_t=\langle\w_{P,t},\x_t^P\rangle\in\mathbb{R}.$
	    \STATE Receive the target $y_t\in\mathbb{R}$, and suffer loss $\ell(p_t,y_t).$
	    \STATE Update $\w_{P,t}$ using~(\ref{equation:update model1}) where $\tau_t=1/\sqrt{t}.$
	    \ENDFOR
	\end{algorithmic}
\end{algorithm}

\subsection{Complete Matrix $M$}
\label{section:complete}
Back to the example of ecosystem protection, each feature is represented by the data gathered by a sensor. In our scenario, when some old sensor disappears, it means that the corresponding feature will vanish forever. In other words, for each row in $M$, the remaining or observed entries are always fewer than or equal to the entries in the preceding row. Besides, each element in the current row is observed only once and the vanishings of features are uniformly at random since the corresponding sensors expire uniformly at random. Thus this setting can be formulated as the sampling each row uniformly at random without replacement in matrix completion problem. Traditional matrix completion methods with nuclear norm minimization~\cite{DBLP:journals/jmlr/Recht11, DBLP:journals/cacm/CandesR12} are not appropriate in our setting because they usually assume that the observed entries are sampled uniformly at random from the whole matrix, whereas in our setting, entries are observed in the certain rule mentioned above. On the other hand, what we handle is data stream, which means it is more natural and appropriate to deal with it sequentially. Thus, it is desirable to complete each row immediately when receiving it, which cannot be resolved by traditional matrix completion approaches neither.

Specifically, for a matrix $K\in\mathbb{R}^{n\times m}$, let $K_{(i)}$ and $K^{(j)}$ denote the $i$th row and $j$th column of $K$, respectively. For a set $\Omega \subset\{1,\ldots,n\}$, the vector $\mathbf{x}_\Omega\in \mathbb{R}^{|\Omega|}$ contains elements of vector $\mathbf{x}$ indexed by $\Omega$. Similarly the matrix $K_\Omega\in \mathbb{R}^{|\Omega|\times m}$ has rows of matrix $K$ indexed by $\Omega$. Let $M=[\mathbf{m}_1,\mathbf{m}_2,\cdots,\mathbf{m}_b]^\top\in\mathbb{R}^{b\times d_1}$ be the matrix to be completed. We observe that matrix $A$ and $M$ share the same feature space and the same column of $A$ and $M$ are data gathered by the same sensor. Thus it is reasonable to assume that matrix $A$ and $M$ are spanned by the same row space. Therefore, we can leverage $A$ to obtain the row space of $M$ and recover each row of $M$. Concretely, to approximate $M$, let $r\leq \min(T_1-b,d_1)$ be the rank of $A$. We calculate the top-$r$ right singular vectors of $A$ denoted by $V=[\mathbf{v}_1,\mathbf{v}_2,\cdots,\mathbf{v}_{r}]$ that is the row space of $A$. Here $r$ is calculated directly rather than being chosen. Since in our online or one-pass setting we can only obtain one instance~(row) at a time, we use Frequent Directions technique~\cite{DBLP:conf/kdd/Liberty13, DBLP:journals/siamcomp/GhashamiLPW16,DBLP:conf/icml/Huang18} to calculate $V$. Frequent Directions can compute row space of a matrix in a streaming way. For each row of $M$ denoted by $\mathbf{m}_i^\top$, we only observe a set $\Omega_{i}$ of $s$ entries denoted by $\mathbf{m}_{i,\Omega_{i}}^\top$. It is equivalent to state that we sample a set $\Omega_{i}$ of $s$ entries uniformly at random without replacement from $\mathbf{m}_i^\top$. We then solve the following optimization problem
\begin{equation}
\label{opt}
\min_{\mathbf{z}\in\R^r}\frac{1}{2}\|\mathbf{m}_{i,\Omega_{i}}-V_{\Omega_{i}}\mathbf{z}\|_2^2
\end{equation}
to recover this row by $\mathbf{m}_i=V\mathbf{z}_\ast$, where $\mathbf{z}_\ast$ is the optimal solution and $V_{\Omega_{i}}$ is the selected columns of $V$ indexed by $\Omega_i$. Since this problem has a closed-form solution 
$\mathbf{z}_\ast=(V_{\Omega_{i}}^\top V_{\Omega_{i}})^{-1}V_{\Omega_{i}}^\top\mathbf{m}_{i,\Omega_{i}},$ we have 
\[
\mathbf{m}_i=V(V_{\Omega_{i}}^\top V_{\Omega_{i}})^{-1} V_{\Omega_{i}}^\top\mathbf{m}_{i,\Omega_{i}}.
\]
The above procedures are summarized in Algorithm~\ref{alg:complete}. Although the algorithm is simple, its proving on the least number of observed entries faces a big challenge, i.e., the entries of the matrix $M$ are observed uniformly at random {\it without} replacement. We leverage a new technique named ``Matrix Chernoff''~\cite{DBLP:journals/aada/Tropp11} to tackle this problem~(detailed proof can be found in the supplementary file). Note that we do not assume that the first row of matrix $M$ must be complete (e.g., in the ecosystem protection example, some sensors may expire before the replacement, rendering the first row of $M$ incomplete). Thus our method is not affected by this situation.

\begin{algorithm}[t]
\caption{Complete Matrix $M$}
\label{alg:complete}
\begin{algorithmic}[1]
\STATE \textbf{Input:} number of observed entries per row, $s$.
\STATE Calculate the top-$r$ right singular vectors of $A$ denoted by $V=[V^{(1)},V^{(2)},\cdots,V^{(r)}]$ by Frequent Directions.
\FOR{$t = T_1-b+1,\ldots,T_1$}
\STATE Sample a set $\Omega_{i}$ of $s$ entries uniformly at random without replacement denoted by $\mathbf{m}_{i,\Omega_{i}}^\top$.
\STATE Calculate $\mathbf{m}_i =
V(V_{\Omega_i}^\top V_{\Omega_i})^{-1}V_{\Omega_i}^\top\mathbf{m}_{i,\Omega_i}$.
\ENDFOR
\STATE \textbf{Output: $M = [\mathbf{m}_1,\cdots,\mathbf{m}_n]^\top$}.
\end{algorithmic}
\end{algorithm}

Let $r\in[\min(T_1-b,d_1)],$ let $U = [\mathbf{u}_1,\mathbf{u}_2,\cdots,\mathbf{u}_r]\in\mathbb{R}^{(T_1-b)\times r}$ and $V = [\mathbf{v}_1,\mathbf{v}_2,\cdots,\mathbf{v}_r]\in\mathbb{R}^{d_1\times r},$ where $\{\mathbf{u}_i\}_1^{r}$ and $\{\mathbf{v}_i\}_1^{r}$ are the top-$r$ left and right singular vectors of $A$.
The incoherence measure for $U$ and $V$ is defined as
\[
\mu(r)=\max\left(\max\limits_{i\in[T_1-b]}\frac{T_1-b}{r}\|U_{(i)}\|^2_2,\max\limits_{i\in[d_1]}\frac{d_1}{r}\|V_{(i)}\|_2^2\right).
\]
The following theorem demonstrates that in the low-rank case where $\rk(A)= r$ when observing \[s\geq7\mu(r)r\ln(2rn/\delta)\] entries, we can recover $M$ exactly with high probability.
\begin{thm}
\label{thm:complete}
Assume the rank of $A$ is $r$, and the number of observed entries in $M_{(i)}$ is $s\geq7\mu(r)r\ln(rb/\delta)$. With a probability at least $1-\delta$, Algorithm \ref{alg:complete} recovers $M_{(i)}$ exactly.
\end{thm}

\begin{remark} We know that there will be fewer and fewer entries in each row as time goes on. Thus we can recover $M$ exactly if only we guarantee that the number of entries in the last row is larger than $7\mu(r)r\ln(rb/\delta)$. For those rows whose entries are fewer than this amount, we simply discard them. Then an intact overlapping period can be used to learn a mapping. Suppose the number of rows that contain entries more than $s$ is $b$, and the column number is $d_1$, then with the free row space of $A$, the sample complexity is only $\Omega(br\ln r)$ which is much smaller than $\Omega(rd_1\ln^2d_1)$ of the conventional matrix completion~\cite{DBLP:journals/jmlr/Recht11}.
\end{remark}

\subsection{Learn Mapping from $N$ to $M$}
There are several methods to learn a relationship between two sets of features including multivariate regression~\cite{DBLP:journals/technometrics/Kibria07}, streaming multi-label learning~\cite{DBLP:journals/jmlr/ReadBHP11}, etc. We follow FESL~\cite{DBLP:conf/nips/Hou0Z17} and choose to use the popular and effective method --- least squares~\cite{stigler1981gauss} which can be formulated as follows:
\[
\min_{\psi:\mathbb{R}^{d_2}\rightarrow\mathbb{R}^{d_1}}\sum\nolimits_{t=T_1-b+1}^{T_1}\frac{1}{2}\|\x_t^P-\psi(\x_t^C)\|_2^2.
\]
If the overlapping period is very short, it is unrealistic to learn a complex relationship between the two spaces due to under-fitting. Instead, we can use a linear mapping to approximate $\psi$. Assume the coefficient matrix of the linear mapping is $\P$, then during rounds $T_1-b+1,\ldots,T_1$, the estimation of $\P$ can be based on linear least square method
\[
\min_{\P\in\mathbb{R}^{d_2\times d_1}}\sum\nolimits_{t=T_1-b+1}^{T_1}\frac{1}{2}\|\x_t^P-\P^\top\x_t^C\|_2^2.
\]
The optimal solution $\P_*$ to the above problem is given by
\[
\P_*=\left(\sum_{t=T_1-b+1}^{T_1}\x_t^C{\x_t^C}^\top\right)^{-1}\left(\sum_{t=T_1-b+1}^{T_1}\x_t^C{\x_t^P}^\top\right).
\]
Note that we do not need a budget to store instances from the overlapping period because during the period from $T_1-b+1$ to $T_1$, $\P_*$ can be calculated in an online way, i.e. we first iteratively calculate $P_1$ and $P_2$,
\[
P_1=P_1+\x_t^C{\x_t^C}^\top\,\text{and}\,P_2=P_2+\x_t^C{\x_t^P}^\top,
\]
then, 
\begin{equation}
\label{equation:linear least square}
\P_*=P_1^{-1}P_2.
\end{equation}
Then if we only observe an instance $\x_t^C\in\mathbb{R}^{d_2}$ from the current feature space, we can recover an instance in the previous feature space by $\psi(\x^C)\in\mathbb{R}^{d_1}$, to which $\w_{P,T_1}$ can be applied.

\subsection{Prediction in Current Feature Space}
\label{section:predict}

From round $t>T_1$, if we keep on updating $\w_{P,t}$ using the recovered data $\psi(\x_t^C)$, i.e.,
\begin{equation}
\label{equation:update model2}
\w_{P,t+1}=\Pi_{\O_P}\left(\w_{P,t}-\tau_t\nabla\ell(\w_{P,t}^\top(\psi(\x_t^{C})),y_t)\right),
\end{equation}
where $\tau_t$ is a varied step size, the learner can mainly calculate two base predictions: $\w_{P,t}^\top(\psi(\x_t^C))$ and $\w_{C,t}^\top\x_t^C$ based on models $\w_{P,t}$ and $\w_{C,t}.$ Through ensembling the base predictions in each round by weighted combination~\cite{DBLP:journals/fcsc/DongYCSM20}, our model is able to follow the best base model theoretically and empirically. We borrow the idea of learning with expert~\cite{DBLP:conf/colt/LuoS15} to realize it. The main idea is to leverage the loss of the last round to adaptively adjust the weight of each base prediction in the next round.

\begin{algorithm}[t]
\caption{Prediction in Current Feature Space}
\label{alg:prediction}
\begin{algorithmic}[1]
\STATE Let $R_{i,T_1}=0,S_{i,T_1}=0,\,i=1,2$.
\FOR{$t = T_1+1,\ldots,T_1+T_2$}
\STATE Predict the weight of each base model $\alpha_{i,t}$ using (\ref{alpha}).
\STATE Receive $\x_t^C\in\mathbb{R}^{d_2}$ and make prediction $p_{1,t}=\w_{P,t}^\top\psi(\x_t^C)\in\mathbb{R}$ and $p_{2,t}=\w_{C,t}^\top\x_t^C\in\mathbb{R}.$
\STATE Calculate our prediction by $\hat{p}_t=\bm{\alpha}_t^\top\bm{p}_t$ where $\bm{p}_t=(p_{1,t},p_{2,t})^\top$.
\STATE Receive $y_t$, each base model suffers loss $\ell_{i,t}=\ell(p_{i,t},y_t)$ and our model suffers $\hat{\ell}_t=\ell(\hat{p}_t,y_t).$ 
\STATE Set $r_{i,t}=\hat{\ell}_t-\ell_{i,t},\ R_{i,t}=R_{i,t-1}+r_{i,t},\ S_{i,t}=S_{i,t-1}+|r_{i,t}|.$
\STATE Update $\w_{P,t}$ and $\w_{C,t}$ using~(\ref{equation:update model2}) and~(\ref{equation:update model3}) respectively where $\tau_t=1/\sqrt{t-T_1}$. 
\ENDFOR
\end{algorithmic}
\end{algorithm}

We first give some notations that we need to use here.  
In our case, the number of base models is $2$ where the first base model is $\w_{P,t}$ and the second one is $\w_{C,t}$. It is easy to extend the amount to a positive integer $N>2$ when newer feature spaces appear. To be general, we use $N$ as the number of the base models in the following. Let $\alpha_{i,t},\,i=1,\ldots,N$ be the weight of the $i$th model at time $t$. $\ell_{i,t}\in[0,1]$ is the loss of the $i$th base model at time $t$. Then our prediction $\hat{p}_t$ at time $t$ is the weighted combination of the $N$ base predictions, namely, 
\[\hat{p}_t=\bm{\alpha}_t^\top\bm{p}_t,\]
where $\bm{\alpha}_t=(\alpha_{1,t},\ldots,\alpha_{N,t})^\top$ and $\bm{p}_t=(p_{1,t},\ldots,p_{N,t})^\top$ are the vector of $N$ weights and $N$ base predictions.
We let \[r_{i,t}=\hat{\ell}_t-\ell_{i,t},\, R_{i,t}=\sum_{k=T_1+1}^t r_{i,k},\, S_{i,t}=\sum_{k=T_1+1}^t |r_{i,k}|\] and use $\Delta_N$ to denote the simplex of all distributions over $\{1,\ldots,N\}$. We define the weight function: 
\[w(R,S)=\frac{1}{2}(\Phi(R+1,S+1)-\Phi(R-1,S+1)),\]
where  
$\Phi(R,S)=\exp\left(\max\{0,R\}^2/(3S)\right)$ is the \emph{potential function} with $\Phi(0,0)$ preset to 1.
Then at each round, we set $\alpha_{i,t}$ to be proportional to $w(R_{i,t-1},S_{i,t-1})$:
\begin{equation}
\label{alpha}
\alpha_{i,t}\propto \I_{i,t}w(R_{i,t-1},S_{i,t-1}),
\end{equation} 
where $\I_{i,t}\in[0,1]$ is the confidence of the $i$th base model at time $t$. When receiving instance from the current feature space $\x_t^C\in\mathbb{R}^{d_1}$, we can make prediction \[p_{1,t}=\w_{P,t}^\top\psi(\x_t^C)\in\mathbb{R}\quad \text{and} \quad p_{2,t}=\w_{C,t}^\top\x_t^C\in\mathbb{R}.\] Then with $\bm{\alpha_t}$, we calculate our prediction by $\hat{p}_t=\bm{\alpha}_t^\top\bm{p}_t$. After receiving target $y_t$, our model and the base models suffer loss $\hat{\ell}_t=\ell(\hat{p}_t,y_t)$ and $\ell_{i,t}=\ell(p_{i,t},y_t)$, respectively. Then we update $\w_{C,t}$ by
\begin{equation}
\label{equation:update model3}
\w_{C,t+1}=\Pi_{\O_C}\left(\w_{C,t}-\tau_t\nabla\ell(\w_{C,t}^\top\x_t^{C},y_t)\right),
\end{equation}
and $\w_{P,t}$ by (\ref{equation:update model2}), where $\tau_t$ is a varied step size and $\O_C\subseteq \mathbb{R}^{d_2}$ is the set of linear models in the current feature space. The procedure of learning model in the current feature space is summarized in Algorithm~\ref{alg:prediction} where $i=1,2$ for simplicity.

In the following, we give a theoretical guarantee that we are able to follow the best models by this strategy of weights adjusting. 
We denote the cumulative loss of each base model $i$ in $T_1+1,\ldots,T_1+T_2$ by 
\[L_{i,T_2}=\sum\nolimits_{t=T_1+1}^{T_1+T_2}\ell_{i,t},\,i=1,\ldots,N.\] The cumulative loss of our model in $T_1+1,\ldots,T_1+T_2$ is denoted by
\[
\hat{L}_{T_2}=\sum\nolimits_{t=T_1+1}^{T_1+T_2}\hat{\ell}_t.
\]
Then we have the following theorem.
\begin{thm}
\label{thm:prediction}
For any $\bm{u}\in\Delta_{N_{T_2}}$, the cumulative loss of our model is bounded as follows:
\begin{equation}
\label{single-bound}
\begin{split}
&\hat{L}_{T_2}\\
&\leq \bm{u}^\top\bm{L}_{T_2}+\sqrt{3(\bm{u}\cdot\bm{S}_{T_2})(\ln N_{T_2}+\ln B+\ln(1+\ln N_{T_2}))}\\
&=\bm{u}^\top\bm{L}_{T_2}+\hat{O}(\sqrt{(\bm{u}\cdot\bm{S}_{T_2})\ln N_{T_2}}),
\end{split}
\end{equation}
where $N_{T_2}$ is the total number of the base models created from $T_1+1$ to $T_2$, 
$B=1+\frac{3}{2}\sum_{i=1}^{N_{T_2}} \frac{1}{N_{T_2}}(1+\ln(1+S_{i,T_2}))\leq\frac{5}{2}+\frac{3}{2}\ln(1+T_2),
$ 
$\bm{L}_{T_2}=(L_{1,T_2},\ldots,L_{N,T_2})^\top,\bm{S}_{T_2}=(S_{1,T_2},\ldots,S_{N,T_2})^\top$
 and
  $S_{i,T_2}=\sum_{k=T_1+1}^{T_1+T_2} |r_{i,k}|$ is redefined here for simplicity. We use $\hat{O}$ to hide the ``$\ln\ln$'' terms since they are very small, and thus we consider these terms to be nearly constant.
\end{thm}
\begin{remark}
\label{remark:prediction}
This theorem (proof deferred to supplementary file) shows that our model is comparable to any linear combination of base models. Furthermore, $S_{i,T_2}\leq T_2$ since $S_{i,T_2}$ is the cumulative magnitude of $r_{i,t}\leq 1$. Thus $\bm{u}\cdot\bm{S}_{T_2}\leq T_2.$ If $\bm{u}$ concentrates on the best model with minimum cumulative loss, then the upper bound will become $\hat{L}_{T_2}\leq \min(\bm{L}_{T_2})+\hat{O}(\sqrt{T_2\ln N_{T_2}})$ which is exactly the bound in FESL~\cite{DBLP:conf/nips/Hou0Z17}, which means that our model is comparable to the best model. Yet our bound has several merits over FESL. First, ours is parameter free, which means we do not have to tune $\eta$ that appears in the exponential formula in FESL. Second, $\bm{u}\cdot\bm{S}_{T_2}=T_2$ is the worst case. As long as $r_{i,t},\forall i\in{1,\ldots,N_{T_2}}$ is not always the worst, $\bm{u}\cdot\bm{S}_{T_2}$ will be much smaller than $T_2$. Besides we can utilize any number of base models while in FESL they only focus on two. Note that $\I_{i,t}\in[0,1]$ is the confidence of the $i$th base model at time $t$. We focus on the case when $\I_{i,t}\in\{0,1\}$, which means either the base model participates in our prediction or not. If $\I_{i,t}=0$, it means the $i$th base model is ``asleep'' at round $t$. A base model that has never appeared before should be thought of being asleep for all previous rounds. Thus if the current feature space vanishes and new feature space appears, it means new base models appear and these base models in new feature space can be regarded as being asleep in the current and previous feature space. In this way, we do not need to decide manually which base model should be incorporated or discarded.
\end{remark}

\section{Related Works}
\label{section:Related Work}
Table~\ref{table:related works} exhibits all the works related to ours. In the following, we introduce these works and discuss their differences.

Our work is most related to FESL~\cite{DBLP:conf/nips/Hou0Z17}. It proposes a setting called ``feature evolvable streaming learning''. The authors observe that in learning with data streams, old features can vanish and new ones can occur. To make the problem tractable, they assume there is an overlapping period that contains samples from both feature spaces. Then, they learn a mapping from new features to old features, and in this way both the new and old models can be used for prediction. The overlapping period comes from the assumption that the old features vanish simultaneously. However, usually this assumption does not hold. A more practical assumption is that different features could vanish unpredictably and thus there will be no intact overlapping period. In this paper, we focus on this new setting and propose an effective method to tackle it. 

Another very related work is OPID~\cite{DBLP:journals/pami/HouZ18}, which also handles evolving features in data streams. In this scenario, when old features vanish, part of them survive and continue existing with the  emerging features. These surviving features are called \emph{overlapping features} because they are like the overlapping of the old and new feature spaces in the features dimension. Since its scenario is different from ours, the technical challenges and solutions are also different. Similar to OPID~\cite{DBLP:journals/pami/HouZ18}, REFORM~\cite{DBLP:conf/icml/YeZ0Z18} also assumes that there are overlapping features when feature evolves. It uses optimal transport to learn mapping from the two different feature spaces. Besides it does not consider streaming mode but batch one. 

\begin{table}[!t]
\renewcommand\arraystretch{1.2}
\caption{\small Summary of related works.} 
\label{table:related works}
\centering
\small
\setlength{\tabcolsep}{1mm}
\begin{tabular}{c|c} 
\hline
\makecell[c]{Category} & \makecell[c]{Related Works} \\
\hline
\makecell[c]{Feature Evolvable \\Streaming Learning} & \makecell[c]{FESL~\cite{DBLP:conf/nips/Hou0Z17} (most related), OPID~\cite{DBLP:journals/pami/HouZ18}, \\REFORM~\cite{DBLP:conf/icml/YeZ0Z18},  Zhang et al.~\cite{DBLP:conf/icdm/ZhangZLDZW15, DBLP:journals/tkde/ZhangZL0ZW16}, \\Beyazit el al.~\cite{DBLP:conf/aaai/BeyazitA019}, He et al.~\cite{DBLP:conf/ijcai/HeWWBC019}, SF$^2$EL~\cite{DBLP:journals/corr/abs-2007-11280}} \\
\hline
\makecell[c]{Learning \\with Streams} & \makecell[c]{evolving neural networks~\cite{DBLP:conf/ijcnn/LeiteCG09}, \\core vector machines~\cite{DBLP:conf/icml/TsangKK07}, \\$k$-nearest neighbour~\cite{DBLP:journals/tkde/AggarwalHWY06}, \\online bagging \& boosting~\cite{DBLP:conf/smc/Oza05}, \\ weighted ensemble classifiers~\cite{DBLP:conf/kdd/WangFYH03,DBLP:conf/pakdd/NguyenWNW12},\\online learning~\cite{DBLP:conf/icml/Zinkevich03,DBLP:journals/jmlr/HoiWZ14}}\\
\hline
\makecell[c]{Learning with\\Multiple Features} & \makecell[c]{multi-view learning~\cite{DBLP:conf/aaai/LiJZ14,DBLP:conf/icml/MusleaMK02,DBLP:journals/corr/abs-1304-5634}, \\transfer learning~\cite{DBLP:journals/tkde/PanY10,DBLP:conf/icml/RainaBLPN07},\\online transfer learning~\cite{DBLP:journals/ai/ZhaoHWL14}}\\
\hline
        \end{tabular}
\end{table}

Learning with trapezoidal data streams~\cite{DBLP:conf/icdm/ZhangZLDZW15,DBLP:journals/tkde/ZhangZL0ZW16} is also a closely related work to us. They deal with trapezoidal data stream where the instance and feature can doubly increase. Though their feature space evolves, the setting that new data always have overlapping features with all old data is different from our work. Recently, some works claim that they can tackle the situation where features could vary arbitrarily at different time steps~\cite{DBLP:conf/aaai/BeyazitA019,DBLP:conf/ijcai/HeWWBC019}. However, they still need to assume that there are relations between old and new features. Note that ``no free lunch'' says that if we want the model can generalize on unseen data, there must be relations between the training data and the unseen one~\cite{DBLP:journals/tec/DolpertM97}. Thus the requirement on the relations between the old and new features can be regarded as a new perspective of ``no free lunch'' in learning with feature evolution. In addition, labels may be rarely given when learning with feature evolvable streams and this setting has been studied in \cite{DBLP:journals/corr/abs-2007-11280} recently.

Our work is also related to data stream mining, such as evolving neural networks~\cite{DBLP:conf/ijcnn/LeiteCG09}, core vector machines~\cite{DBLP:conf/icml/TsangKK07}, $k$-nearest neighbour~\cite{DBLP:journals/tkde/AggarwalHWY06}, online bagging \& boosting~\cite{DBLP:conf/smc/Oza05} and weighted ensemble classifiers~\cite{DBLP:conf/kdd/WangFYH03,DBLP:conf/pakdd/NguyenWNW12}. For more details, please refer to an overview on data stream mining~\cite{DBLP:books/daglib/p/Aggarwal10}. These conventional data stream mining methods usually assume that the data samples are described by the same set of features, while in many real streaming tasks feature often changes. 

Online learning~\cite{DBLP:conf/icml/Zinkevich03,DBLP:journals/jmlr/HoiWZ14} is another related topic from the area of machine learning. It can naturally handle the data streams since it assumes that the data come in a streaming way. Specifically, at each round, after the learner makes prediction on the given instance, the adversary will reveal its loss, with which, the learner will make better prediction to minimize the total loss through all rounds. Online learning has been extensively studied under different settings, such as learning with experts~\cite{DBLP:books/daglib/0016248} and online convex optimization~\cite{DBLP:journals/ml/HazanAK07,DBLP:journals/ftml/Shalev-Shwartz12}. There are strong theoretical guarantees for online learning, and it usually uses regret or the number of mistakes to measure the performance of the learning procedure. However, most of existing online learning algorithms are limited to the case that the feature set is fixed.

Other related topics involving multiple feature sets include multi-view learning~\cite{DBLP:conf/aaai/LiJZ14,DBLP:conf/icml/MusleaMK02,DBLP:journals/corr/abs-1304-5634}, transfer learning~\cite{DBLP:journals/tkde/PanY10,DBLP:conf/icml/RainaBLPN07}, etc. Although multi-view learning exploits the relation between different sets of features as ours, there exists a fundamental difference: multi-view learning assumes that every sample is described by multiple feature sets simultaneously, whereas in PUFE only few samples in the feature switching period have two sets of features. Transfer learning usually assumes that data come by batches, few of them consider the streaming cases where data arrives sequentially and cannot be stored completely. One exception is online transfer learning~\cite{DBLP:journals/ai/ZhaoHWL14} in which data from both sets of features arrive sequentially. However, they assume that all the feature spaces must appear simultaneously during the whole learning process while such an assumption is not available in PUFE.

\section{Experiments}
\label{section:experiments}

In this section, we first introduce the datasets that we use. Then we describe the compared approaches and experimental settings. Finally, we present the results of our experiments.

\subsection{Datasets}
We conduct our experiments on $11$ datasets consisting of $9$ synthetic datasets and $2$ real datasets. To generate synthetic data, we randomly choose some datasets from different domains including \emph{economy}, \emph{biology}, \emph{literature}, etc.~\footnote{Datasets can be found in http://archive.ics.uci.edu/ml/.} We artificially map the original datasets into another feature space by random matrices, then we have data both from the previous and current feature space. Then there are relationships between the previous and current feature space which is exactly the ``no free lunch'' that is mentioned in Section~\ref{section:Related Work}. Since the original data are in batch mode, we manually make them come sequentially. In the overlapping period, we discard entries of each row uniformly at random from the remaining features obeying the vanishing rule mentioned in Section~\ref{section:complete}. In this way, synthetic data are completely generated. Besides the $9$ synthetic datasets, we also conduct our experiments on $2$ real datasets that are collected by ourselves. They are ``RFID'' and ``Amazon''. For the real data, the ecosystem protection task mentioned in the Introduction is a good example to demonstrate that the requirement on the evolving feature awareness is necessary. Although we do not collect the real datasets of this exact task, we find other tasks where the feature evolution happens, that are the moving goods detection task~\cite{DBLP:conf/infocom/WangX0XL16} and the products’ quality prediction task in Amazon product-user review datasets~\cite{DBLP:conf/sigir/McAuleyTSH15,DBLP:conf/www/HeM16}. 

``RFID'' contains $450$ instances from the previous and current feature space respectively. The previous feature space has $78$ features while in the current feature space, there are $72$ features. RFID technique is widely used to do moving goods detection~\cite{DBLP:conf/infocom/WangX0XL16}. This dataset uses the RFID technique to gather the location's coordinate of the moving goods attached by RFID tags. Concretely, several RFID aerials are arranged in the indoor area. In each round, RFID aerials received the tag signals. Then the goods with tag moved, at the same time, the goods' coordinate is recorded. Before the aerials expire, new aerials are arranged beside the old ones to avoid the situation where no aerials exist. So in this overlapping period, data are from both the previous and current feature spaces. After the old aerials expired, the new ones continue to receive signals. Then data only from the current feature space remain. The overlapping period in this dataset is complete, so we simulate unpredictable feature evolution like we did on the synthetic data. Therefore, the modified RFID data satisfy our assumptions. This dataset can be found in http://www.lamda.nju.edu.cn/data\_RFID.ashx.

``Amazon'' comes from Amazon product-user review datasets~\cite{DBLP:conf/sigir/McAuleyTSH15,DBLP:conf/www/HeM16} over ``Movies and TV''.~\footnote{http://jmcauley.ucsd.edu/data/amazon/links.html.} Each product is reviewed by several users over several years. The elements of each review are as follows:
\begin{itemize}
\item reviewerID - ID of the reviewer, e.g. A2SUAM1J3GNN3B
\item asin - ID of the product, e.g. 0000013714
\item reviewerName - name of the reviewer
\item helpful - helpfulness rating of the review, e.g. 2/3
\item reviewText - text of the review
\item overall - rating of the product
\item summary - summary of the review
\item unixReviewTime - time of the review (unix time)
\item reviewTime - time of the review (raw)
\end{itemize} 
We want to predict each product's quality from year $2006$ to $2008$ according to the ratings of its users. Therefore, each instance represents a product and each feature of this instance is its users' rating. As time goes on, some users disappear, e.g., they signed out of their accounts, and some new users join. Thus, the features will evolve, which means old features will disappear and new feature will emerge. We find some period where old and new features both exist and make this dataset satisfy our assumption. The label of each product is its quality that is calculated by the weighted combination of each user's rating. The weight of each rating is calculated by the quality of its user and the quality of each user is calculated by the ``helpfulness'' of the user's reviews. The number of instances are $23025$, in the previous feature space, there are $278$ features while in the current feature space, there are $227$ features.

\begin{table*}[!t]
\renewcommand\arraystretch{1.2}
\caption{\small ``-P'' and ``-U'' means predictable and unpredictable scenarios respectively. The first nine big rows (each contains two unit rows) are the accuracy with its standard deviation on synthetic datasets (the larger the better). The last two big rows are the mean square error (MSE) with its standard deviation on real datasets (the smaller the better). The best ones among all the methods are bold. Black dot indicates the best among three base models, i.e., NOGD, ROGD-f, ROGD-u. Dash lines mean FESL cannot work in unpredictable scenario. Note that we do not have to be better than the base models and it is sufficient to be comparable with them. T-tests with 95\% confidence validate that there are no significant difference between our model and the best base models on all datasets except german-P, german-U and RFID-P. On these three datasets, our results are significantly better than the best base models with 95\% confidence.} 
\label{table:accuracy_synthetic}
\centering
\small
\setlength{\tabcolsep}{3.2mm}
\begin{tabular}{l|l|l|l|c|c|l} 
\hline
\makecell[c]{Dataset} & \makecell[c]{NOGD} & \makecell[c]{ROGD-f} & \makecell[c]{ROGD-u} & \makecell[c]{FESL-c} & \makecell[c]{FESL-s} & \makecell[c]{PUFE (Ours)} \\
\hline
australian-P & .8073$\pm$.0037 & .7727$\pm$.0495 & \textbf{.8676$\pm$.0026}$\bullet$ & \textbf{.8676$\pm$.0026} & .8674$\pm$.0026 & \textbf{.8676$\pm$.0026} \\ 
    australian-U & .8073$\pm$.0037 & .7031$\pm$.1042 & .8515$\pm$.0099$\bullet$ & --- & --- & \textbf{.8535$\pm$.0080} \\ 
    \hline
credit-a-P & .7710$\pm$.0057 & .6369$\pm$.0808 & \textbf{.7892$\pm$.0165}$\bullet$ & .7861$\pm$.0154 & .7861$\pm$.0158 & .7884$\pm$.0161 \\ 
    credit-a-U & \textbf{.7710$\pm$.0057}$\bullet$ & .6025$\pm$.1157 & .7052$\pm$.0738 & --- & --- & .7608$\pm$.0198 \\ 
    \hline
credit-g-P & .6862$\pm$.0040 & .6932$\pm$.0279 & \textbf{.7338$\pm$.0059}$\bullet$ & \textbf{.7338$\pm$.0059} & .7337$\pm$.0058 & \textbf{.7338$\pm$.0059} \\ 
    credit-g-U & .6862$\pm$.0040 & .6394$\pm$.0683 & \textbf{.7277$\pm$.0112}$\bullet$ & --- & --- & \textbf{.7277$\pm$.0112} \\ 
    \hline
diabetes-P & .6403$\pm$.0020 & .6281$\pm$.0709 & \textbf{.6795$\pm$.0018}$\bullet$ & .6785$\pm$.0022 & .6786$\pm$.0019 & \textbf{.6795$\pm$.0018} \\ 
    diabetes-U & .6403$\pm$.0020 & .5120$\pm$.1075 & .6620$\pm$.0151$\bullet$ & --- & --- & \textbf{.6625$\pm$.0145} \\ 
    \hline
dna-P & .7471$\pm$.0035 & .6186$\pm$.0988 & .7739$\pm$.0352$\bullet$ & .7739$\pm$.0352 & \textbf{.7745$\pm$.0344} & .7739$\pm$.0352 \\ 
dna-U & .7471$\pm$.0035 & .5886$\pm$.0762 & .7518$\pm$.0323$\bullet$ & --- & --- & \textbf{.7543$\pm$.0286} \\
\hline
german-P & .7024$\pm$.0011$\bullet$ & .6997$\pm$.0002 & .6998$\pm$.0002 & .7027$\pm$.0010 & .7037$\pm$.0009 & \textbf{.7039$\pm$.0009} \\ 
    german-U & .7024$\pm$.0011$\bullet$ & .6869$\pm$.0171 & .6991$\pm$.0031 & --- & --- & \textbf{.7041$\pm$.0008} \\ 
    \hline
kr-vs-kp-P & .6872$\pm$.0024 & .5544$\pm$.0556 & .7451$\pm$.0236$\bullet$ & \textbf{.7469$\pm$.0225} & .7262$\pm$.0166 & .7452$\pm$.0234 \\ 
kr-vs-kp-U & .6872$\pm$.0024 & .5515$\pm$.0343 & .7081$\pm$.0328$\bullet$ & --- & --- & \textbf{.7097$\pm$.0292} \\ 
\hline
splice-P & .6171$\pm$.0009 & .5694$\pm$.0335 & \textbf{.6602$\pm$.0173}$\bullet$ & \textbf{.6602$\pm$.0173} & \textbf{.6602$\pm$.0173} & \textbf{.6602$\pm$.0173} \\ 
splice-U & .6171$\pm$.0009 & .5612$\pm$.0419 & \textbf{.6553$\pm$.0177}$\bullet$ & --- & --- & \textbf{.6553$\pm$.0177} \\ 
\hline
svmguide3-P & .7396$\pm$.0008 & .7376$\pm$.0002 & \textbf{.7467$\pm$.0057}$\bullet$ & .7463$\pm$.0055 & .7465$\pm$.0056 & \textbf{.7467$\pm$.0057} \\ 
svmguide3-U & .7396$\pm$.0008 & .6442$\pm$.0477 & \textbf{.7562$\pm$.0102}$\bullet$ & --- & --- & .7561$\pm$.0103 \\ 
\hline
\hline
RFID-P & 2.175$\pm$0.058 & 1.641$\pm$0.084 & \textbf{1.297$\pm$0.082}$\bullet$ & 1.309$\pm$0.081 & 1.309$\pm$0.082 & 1.300$\pm$0.082 \\ 
    RFID-U & 2.175$\pm$0.058 & 2.177$\pm$0.092 & 1.719$\pm$0.069$\bullet$ & --- & --- & \textbf{1.578$\pm$0.064} \\ 
    \hline
Amazon-P & \textbf{.0060$\pm$.0000}$\bullet$ & .0062$\pm$.0001 & .0061$\pm$.0001 & .0060$\pm$.0001 & .0060$\pm$.0001 & \textbf{.0060$\pm$.0000} \\ 
    Amazon-U & \textbf{.0060$\pm$.0000}$\bullet$ & .0064$\pm$.0001 & .0062$\pm$.0001 & --- & --- & \textbf{.0060$\pm$.0000} \\ 
    \hline
    
        \end{tabular}
\end{table*}

\subsection{Compared Approaches and Settings}
Since we focus on FESL, we compare our PUFE with five baselines which are all introduced in FESL~\cite{DBLP:conf/nips/Hou0Z17}: 
\begin{itemize}
\item NOGD: Abbreviation of ``Naive Online Gradient Descent''. Once the feature space changed, the online gradient descent algorithm will be invoked from scratch.
\item ROGD-f: Abbreviation of ``Fixed Recovered Online Gradient Descent''. It uses the classifier learned from the previous feature space by online gradient descent to do predictions on the recovered data. It does not update itself with the recovered data or in other words, it keeps fixed.
\item ROGD-u: Abbreviation of ``Updating Recovered Online Gradient Descent''. It also utilizes the classifier learned from the previous feature space by online gradient descent to do predictions on the recovered data. It keeps updating itself with the recovered data.
\item FESL-c: One version of FESL~\cite{DBLP:conf/nips/Hou0Z17}. It uses the exponential of loss to update the weight of each base model and combine them with these weights. It has complete overlapping period. 
\item FESL-s: One version of FESL~\cite{DBLP:conf/nips/Hou0Z17}. It also uses the exponential of loss to update the weight of each base model as FESL-c does. Instead of combining all the base models, FESL-s selects the best one. It also has complete overlapping period. 
\end{itemize}

We evaluate the empirical performances of the proposed approaches on classification and regression tasks during rounds $T_1+1,\ldots,T_1+T_2$. We expect the overall performance can be good and not bad at the beginning or any other time step. We first give the accuracy and mean square error over all instances during rounds $T_1+1,\ldots,T_1+T_2$ on synthetic and real datasets, respectively. We conduct experiments on each dataset with two settings, namely, ``predictable'' and ``unpredictable''. In ``predictable'' setting, the overlapping period is predictable and thus intact while in ``unpredictable'' setting, the overlapping period is unpredictable and thus fragmentary. We want to verify though in general ``predictable'' setting, our method can still work well. Besides, ROGD-u, ROGD-f cannot run in ``unpredictable'' setting since they need intact overlapping period. Thus we fill the overlapping period with $0$ to let them work. FESL cannot work in this scenario. Besides, NOGD is not affected by the overlapping period. 

Furthermore, to verify that our model is comparable to the best base model and has good performance when few data are observed, we present the trend of average cumulative loss. Concretely, at each time $t$, the loss $\bar{\ell}_{t}$ of every method is the average of the cumulative loss over $T_1+1,\ldots,t$, namely 
\[ \bar{\ell}_{t}=(1/(t-T_1))\sum\nolimits_{k=T_1+1}^{t}\ell_{k}.\]
The performances of all approaches are obtained by average results over 10 independent runs.
The parameters we need to set are the number of instances in overlapping period, i.e., $b$; the number of instances in previous and current feature space, i.e., $T_1$ and $T_2$; the step size, i.e., $\tau_t$ where $t$ is time. For all baseline methods and our methods, the parameters are the same. In our experiments, we set $b$ to be 10, 20, 25 for synthetic data, 40 for RFID and 50 for Amazon. We set $T_1$ and $T_2$ to be half of the number of instances, and $\tau_t$ to be $1/(c\sqrt{t})$ where $c$ is searched in the range from $10^{-1}$ to $10^2$ with step size $10$.  

\subsection{Results}
\vspace{-0.4em}
The accuracy and mean square error results are shown in Table~\ref{table:accuracy_synthetic}. The first nine big rows (each contains two unit rows) are the accuracy with its standard deviation on synthetic datasets (the larger the better). The last two big rows are the mean square error with its standard deviation on real datasets (the smaller the better). The best ones among all the methods are bold. Black dot indicates the best among three base models, i.e., NOGD, ROGD-f, ROGD-u. Dash lines mean that FESL-c and FESL-s cannot work in the unpredictable scenario. As can be seen, on total 22 cases, our PUFE outperforms other methods on 16 cases. Note that we do not have to be better than the base models (i.e., NOGD, ROGD-f, ROGD-u) and it is sufficient to be comparable with them. To this end, we conduct t-tests with 95\% confidence and the results demonstrate that our model can be comparable with the best base models in most cases (without significant differences) and in german-P, german-U and RFID-P, our model is significantly better than the best base models with 95\% confidence. PUFE also outperforms FESL-c and FESL-s in most cases even in the predictable scenario.

\begin{figure*}[!t]
\centering
\small
\begin{minipage}{0.20\linewidth}\centering
    \vspace{-0cm}
    \includegraphics[width=1\textwidth]{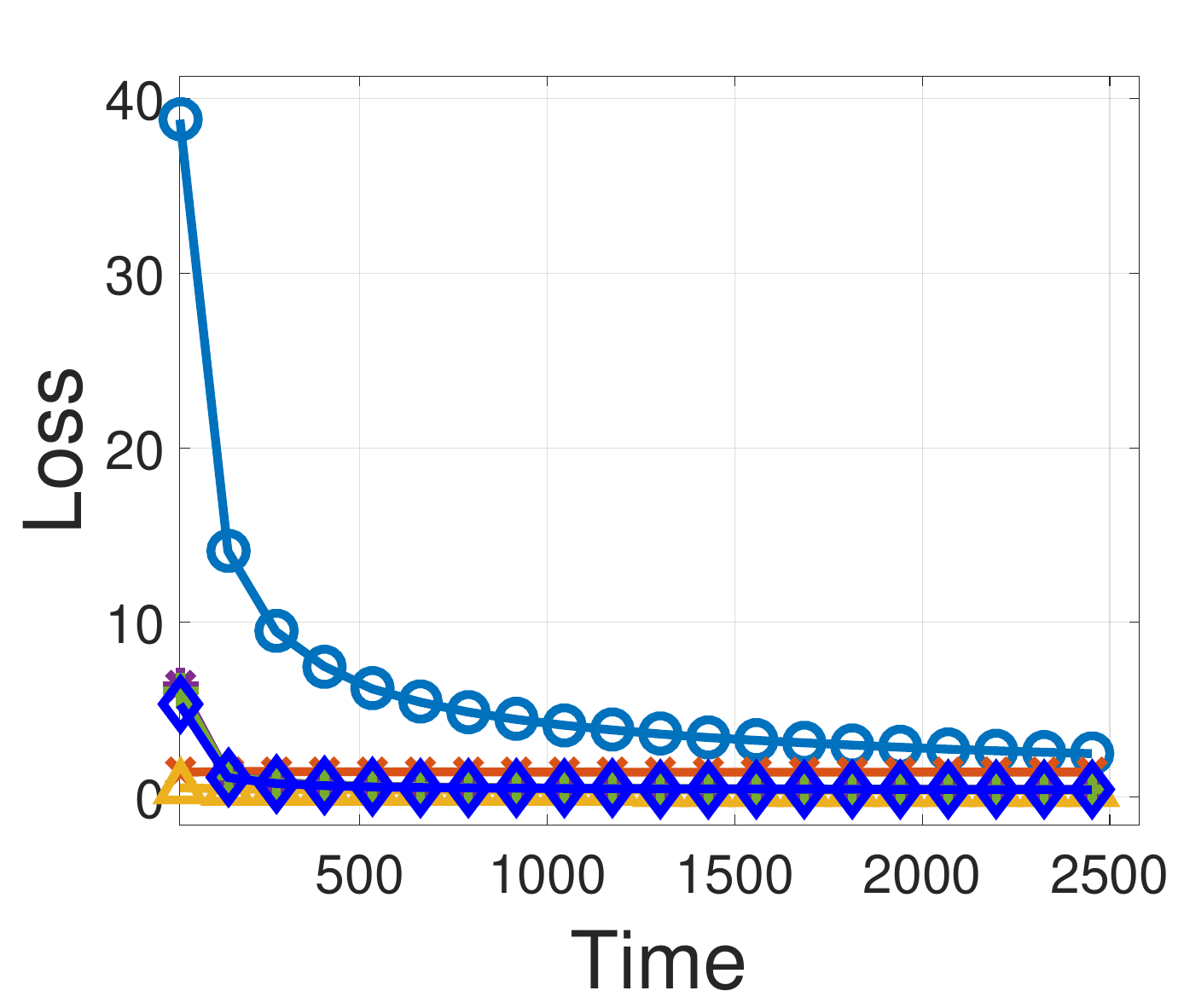}\\
    \vspace{0cm}
    \mbox{\scriptsize\quad\quad(a) \emph{australian}}
\end{minipage}
\begin{minipage}{0.20\linewidth}\centering
    \vspace{-0cm}
    \includegraphics[width=1\textwidth]{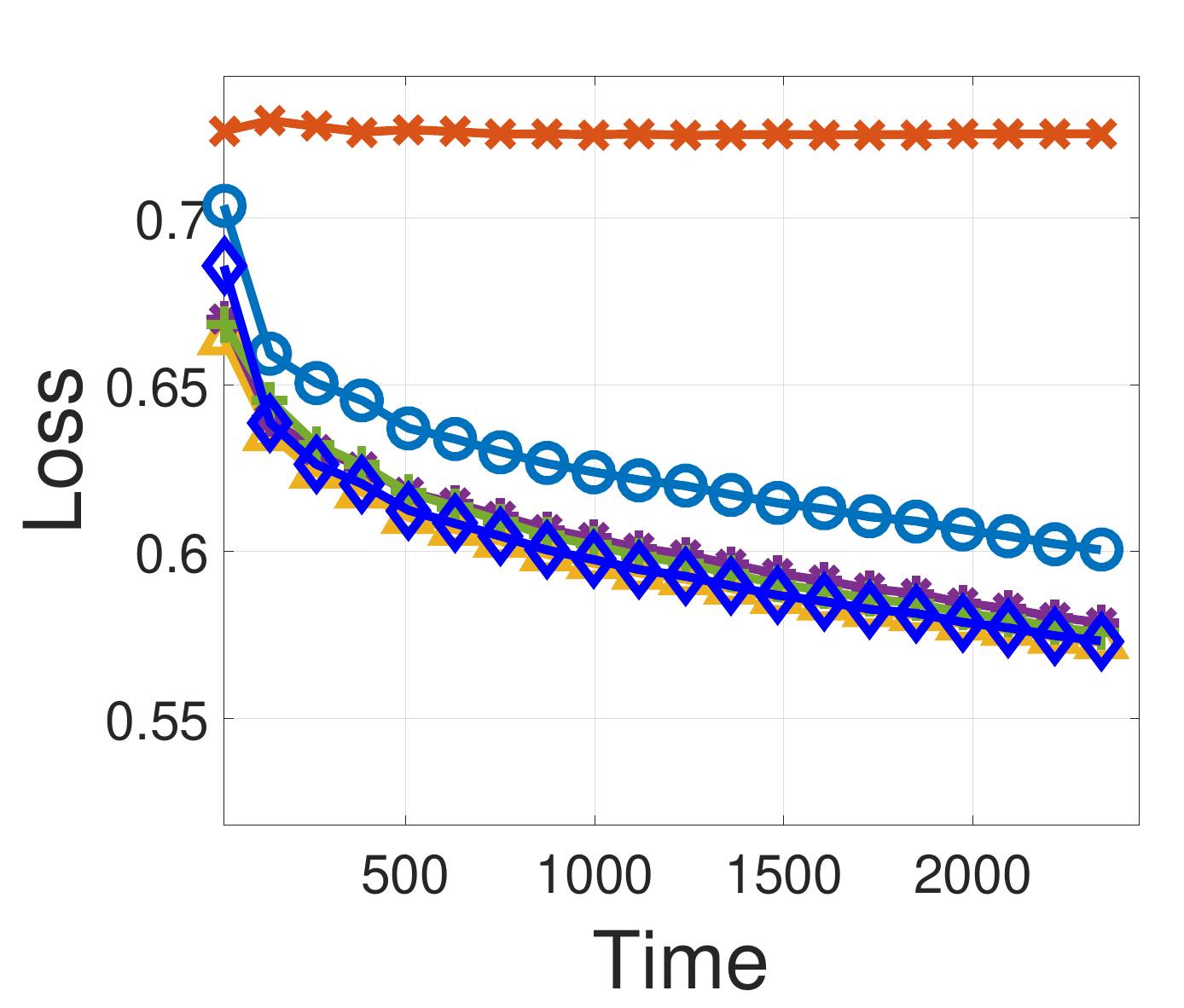}\\
    \vspace{0cm}
    \mbox{\scriptsize\quad\quad(b) \emph{credit-a}}
\end{minipage}
\begin{minipage}{0.20\linewidth}\centering
    \vspace{-0cm}
    \includegraphics[width=1\textwidth]{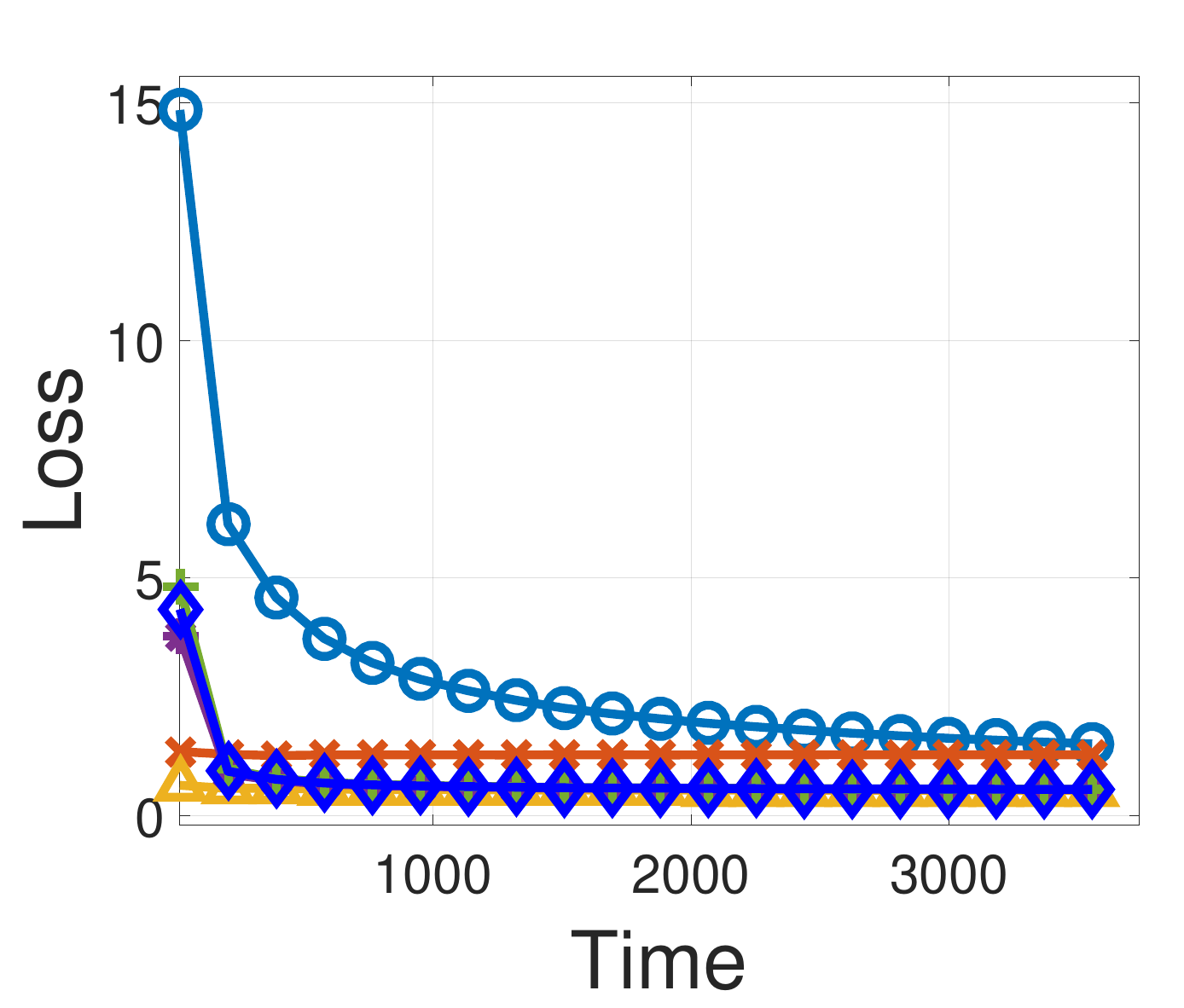}\\
    \vspace{0cm}
    \mbox{\scriptsize\quad\quad(c) \emph{credit-g}}
\end{minipage}
\begin{minipage}{0.20\linewidth}\centering
    \vspace{-0cm}
    \includegraphics[width=1\textwidth]{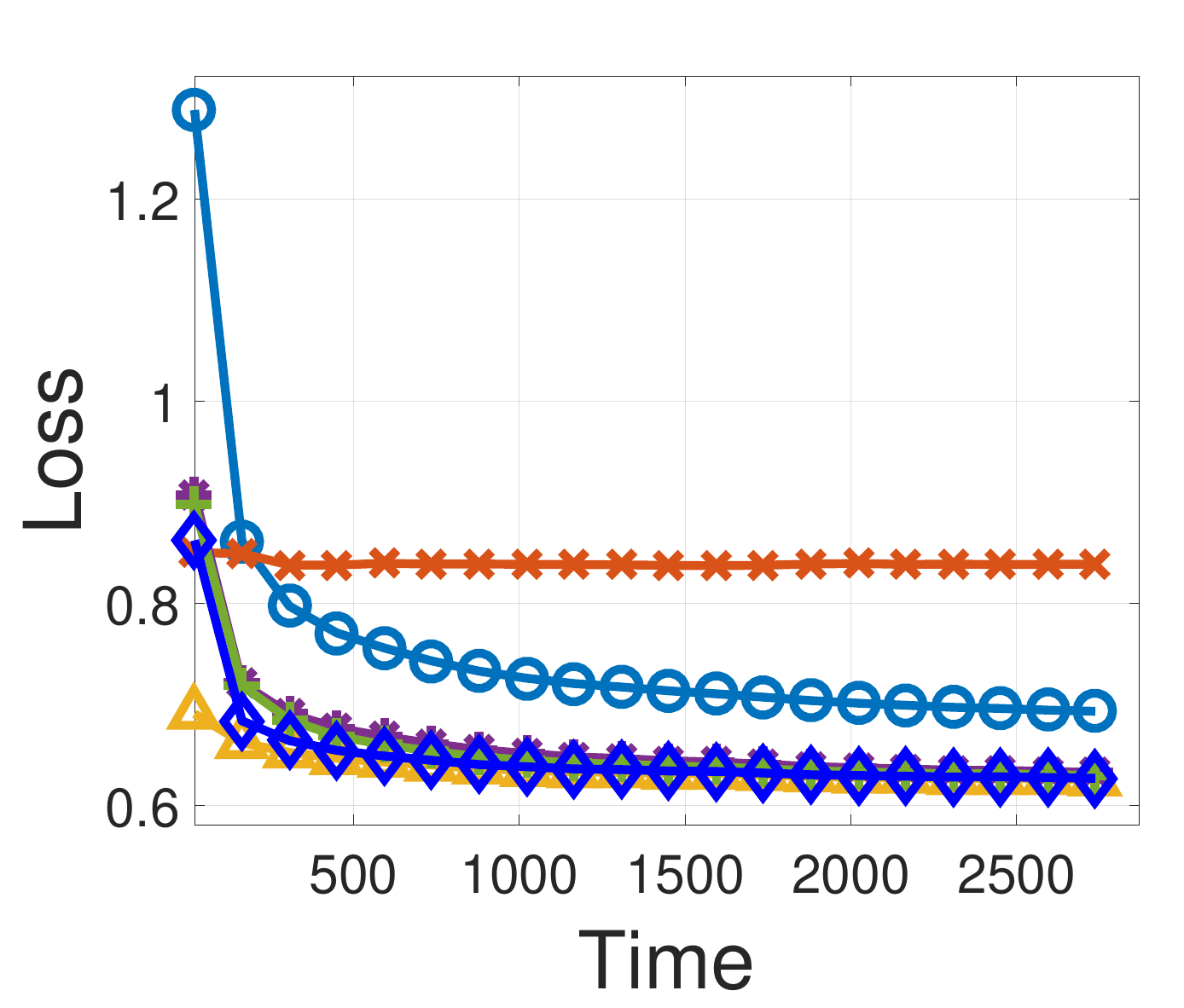}\\
    \vspace{0cm}
    \mbox{\scriptsize\quad\quad(d) \emph{diabetes}}
\end{minipage}

\begin{minipage}{0.20\linewidth}\centering
    \vspace{-0cm}
    \includegraphics[width=1\textwidth]{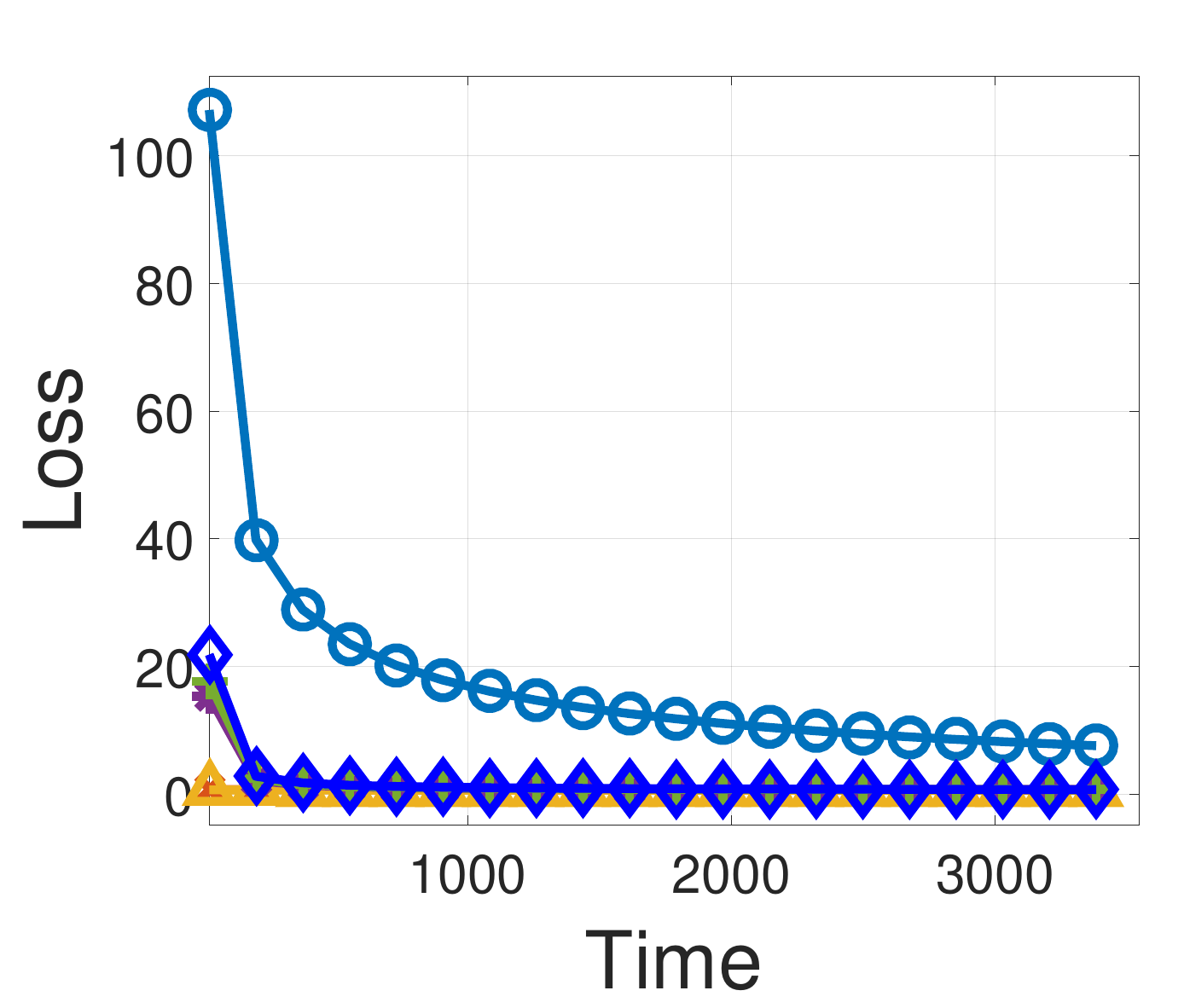}\\
    \vspace{0cm}
    \mbox{\scriptsize\quad\quad(e) \emph{dna}}
\end{minipage}
\begin{minipage}{0.20\linewidth}\centering
    \vspace{0cm}
    \includegraphics[width=1\textwidth]{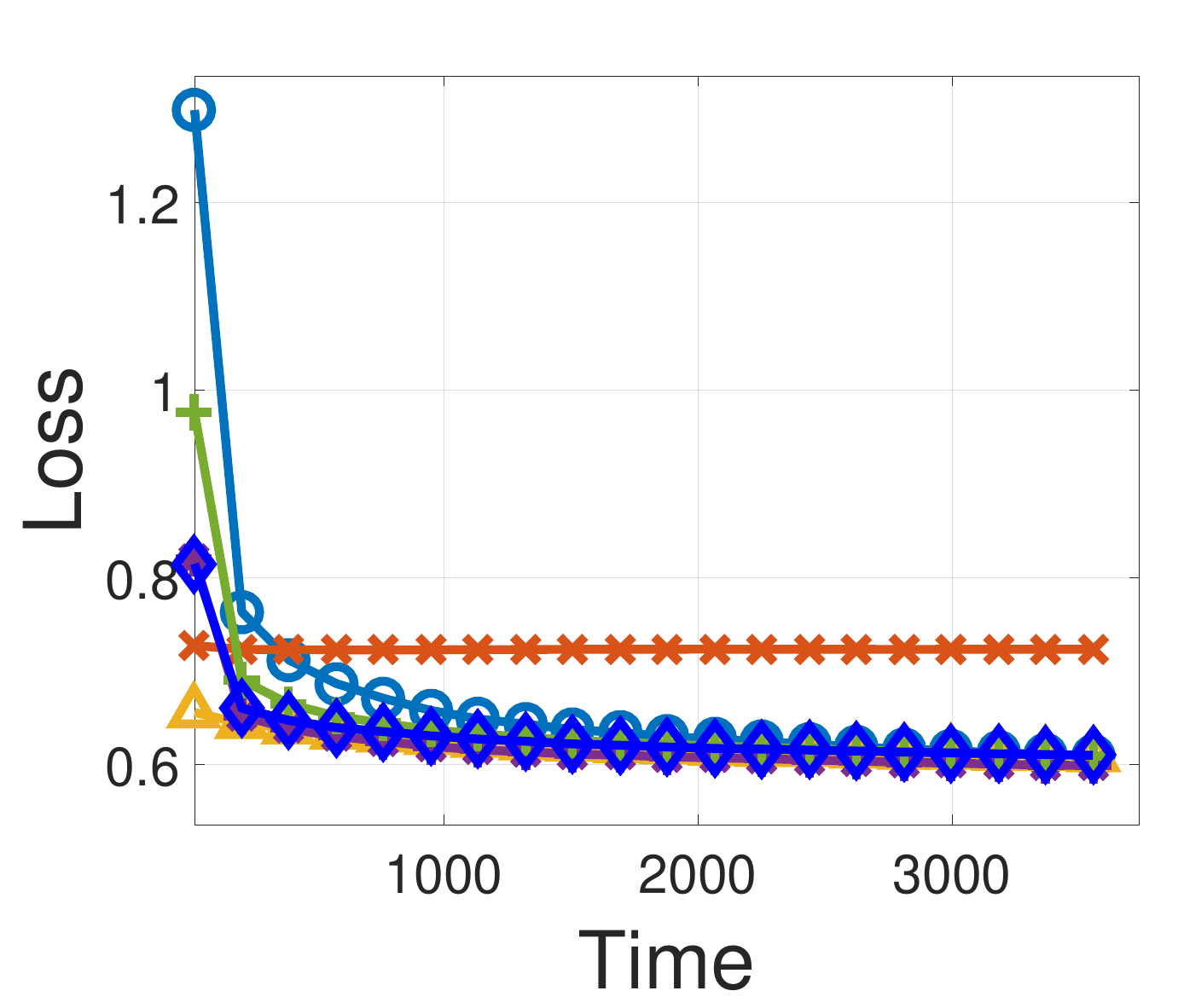}\\
    \vspace{0cm}
    \mbox{\scriptsize\quad\quad(f) \emph{german}}
\end{minipage}
\begin{minipage}{0.20\linewidth}\centering
    \vspace{0cm}    
    \includegraphics[width=1\textwidth]{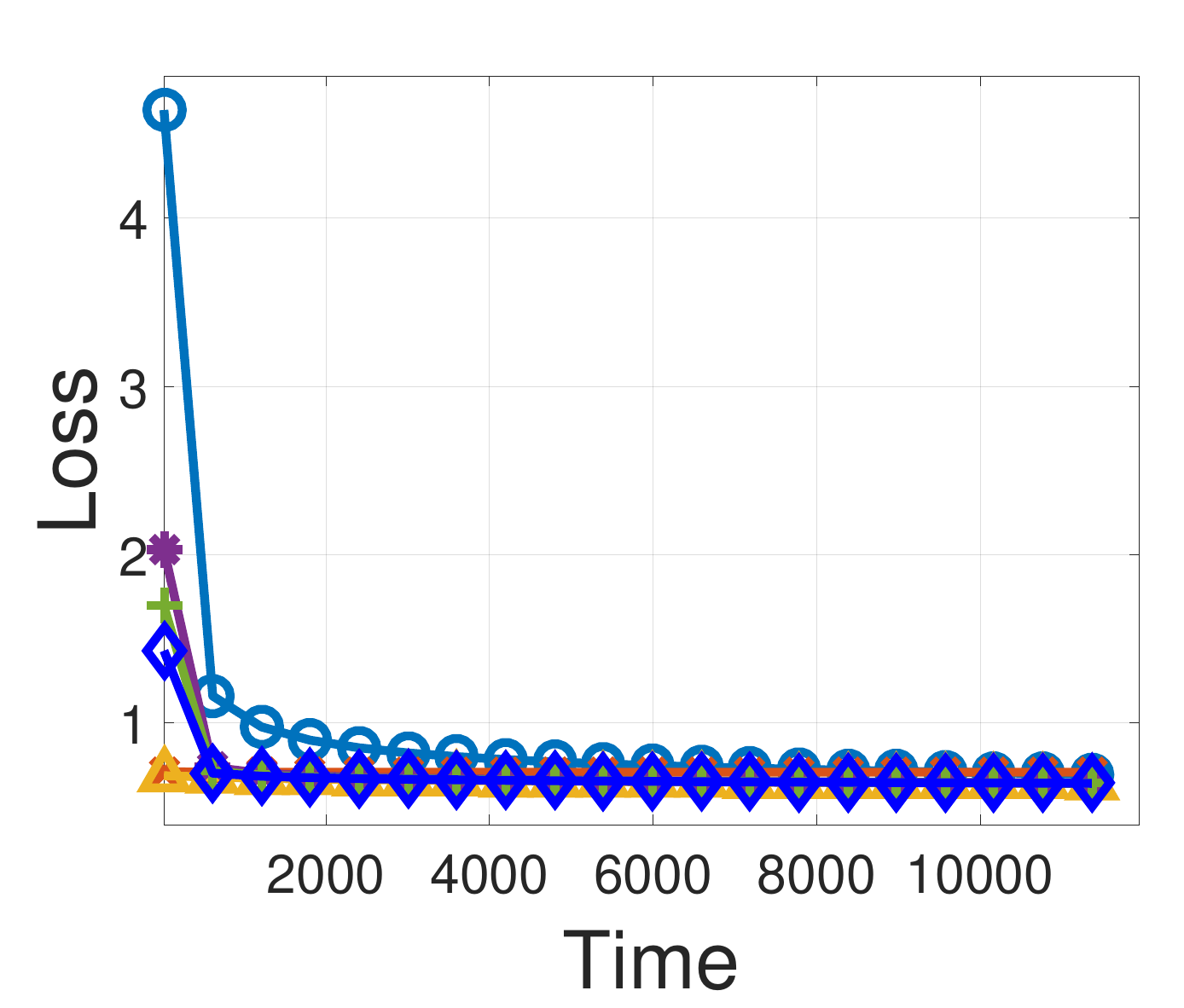}\\
    \vspace{0cm}
    \mbox{\scriptsize\quad\quad(g) \emph{kr-vs-kp}}
\end{minipage}
\begin{minipage}{0.20\linewidth}\centering
    \vspace{0cm}
    \includegraphics[width=1\textwidth]{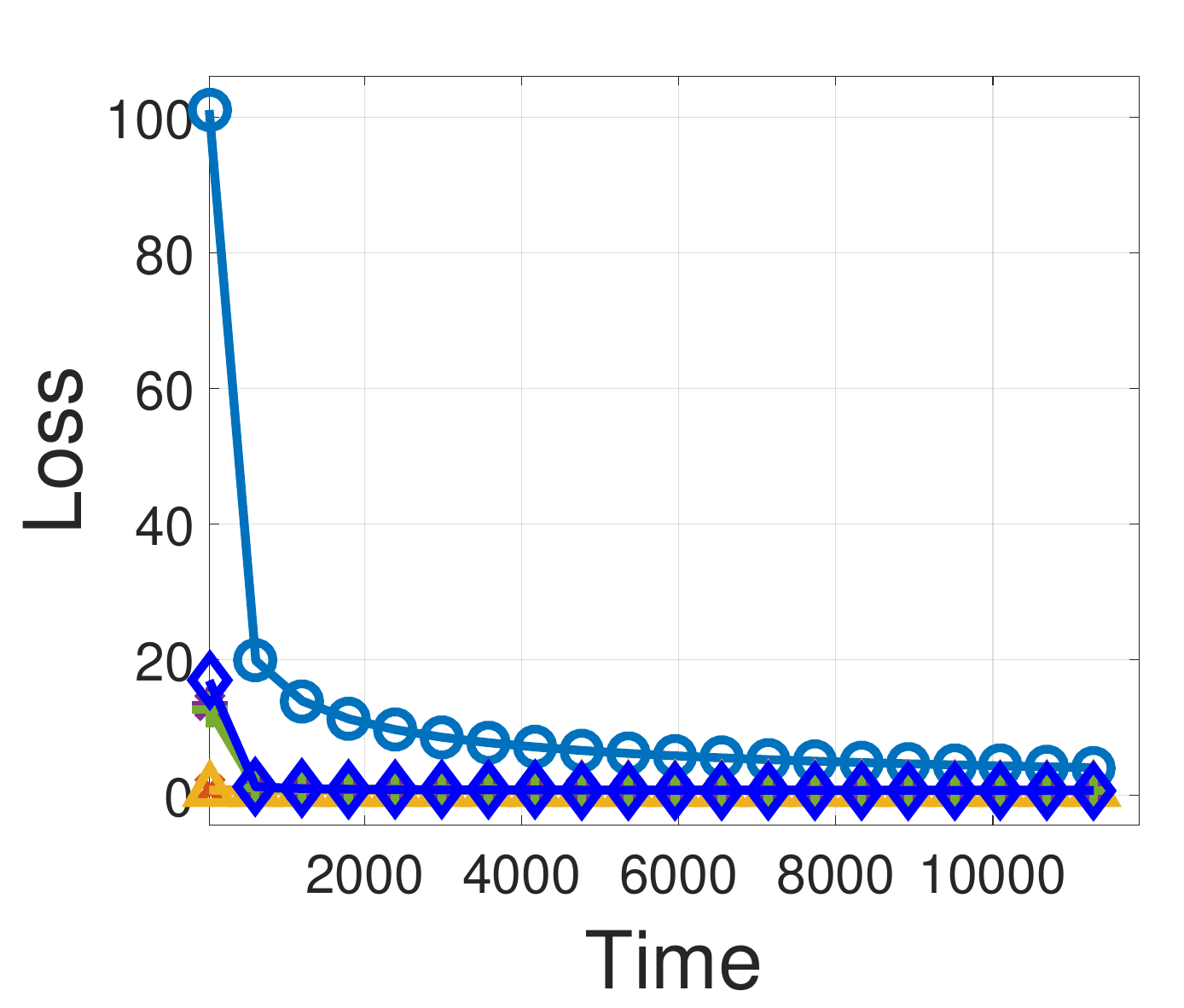}\\
    \vspace{0cm}
    \mbox{\scriptsize\quad\quad(h) \emph{splice}}
\end{minipage}

\begin{minipage}{0.20\linewidth}\centering
    \vspace{0cm}
    \includegraphics[width=1\textwidth]{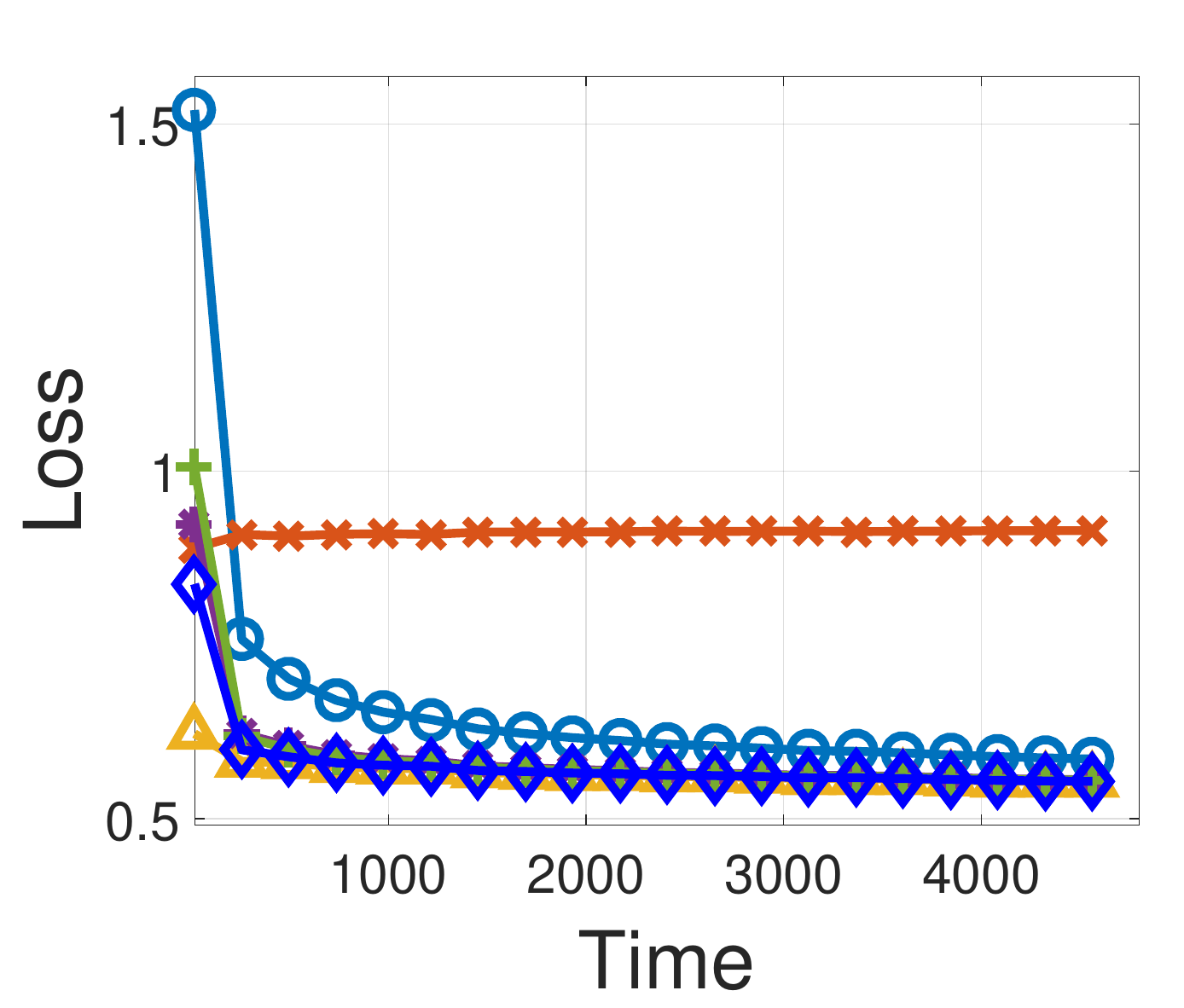}\\
    \vspace{0cm}
    \mbox{\scriptsize\quad\quad(i) \emph{svmguide3}}
\end{minipage}
\begin{minipage}{0.20\linewidth}\centering
    \vspace{-0cm}
    \includegraphics[width=1\textwidth]{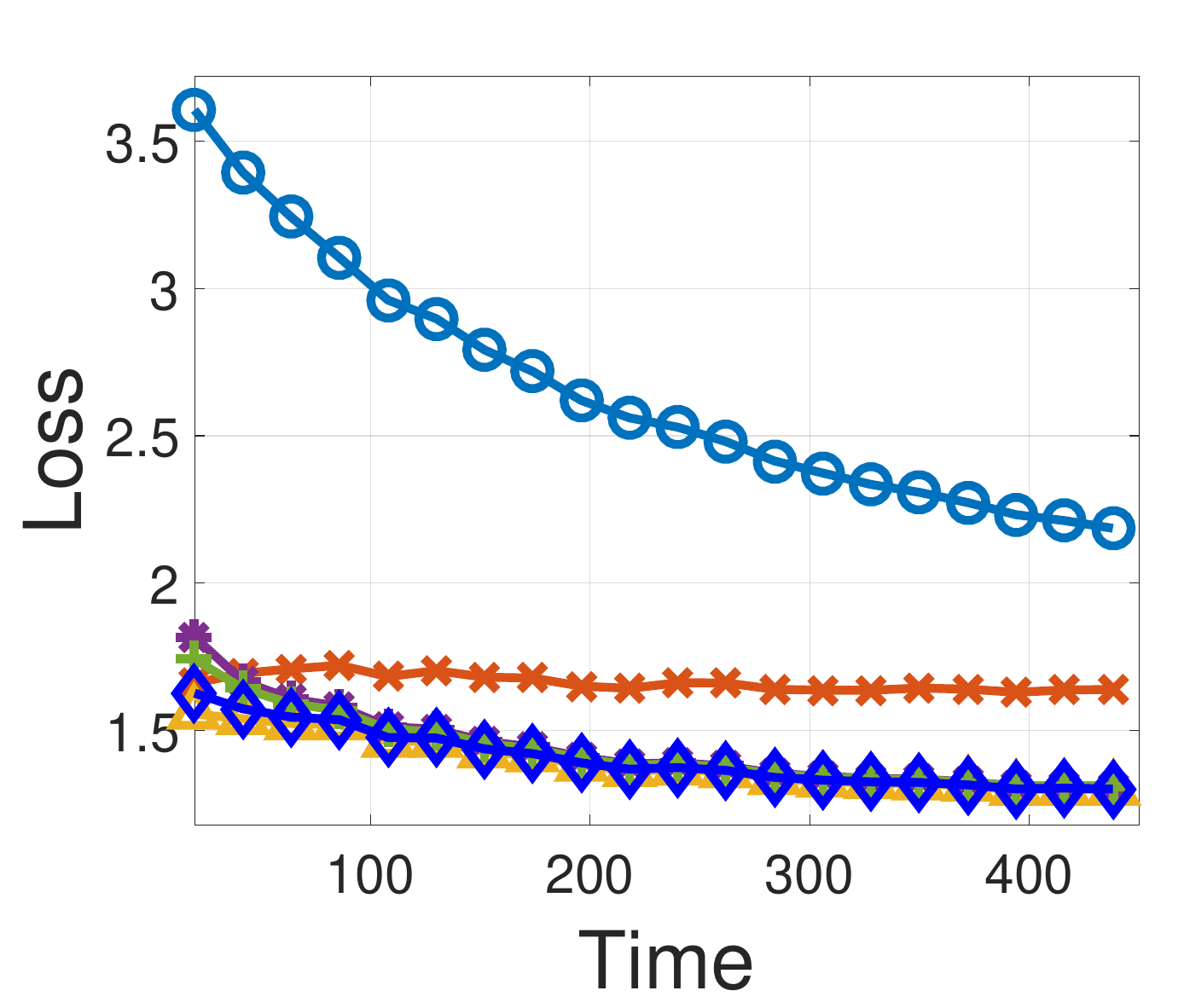}\\
    \vspace{0cm}
    \mbox{\scriptsize\quad\quad(j) \emph{RFID}}
\end{minipage}
\begin{minipage}{0.20\linewidth}\centering
    \vspace{-0cm}
    \includegraphics[width=1\textwidth]{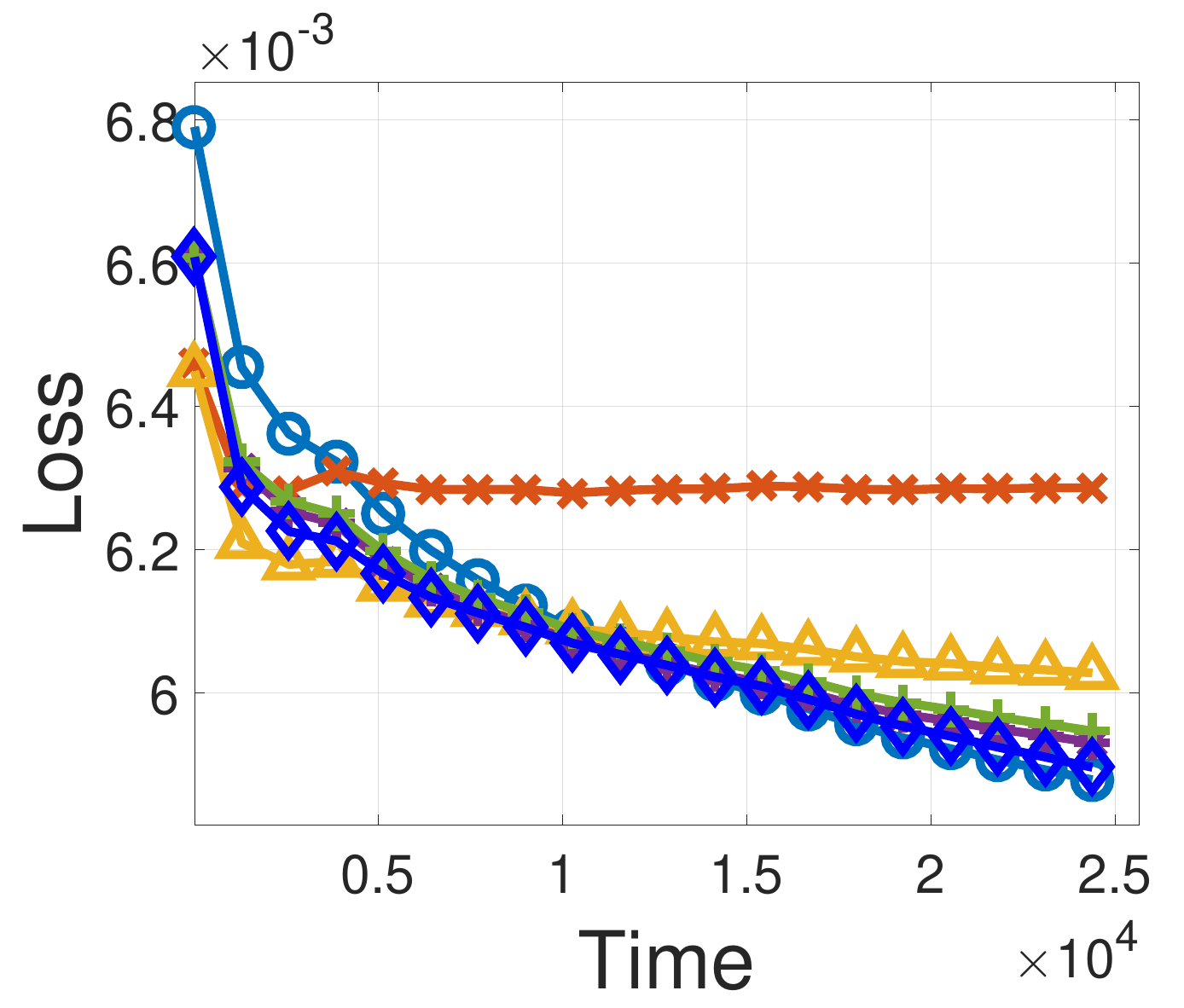}\\
    \vspace{0cm}
    \mbox{\scriptsize\quad\quad(k) \emph{Amazon}}
\end{minipage}
\begin{minipage}{0.20\linewidth}
    \centering
    \vspace{0.1cm}
    \includegraphics[width=0.8\textwidth]{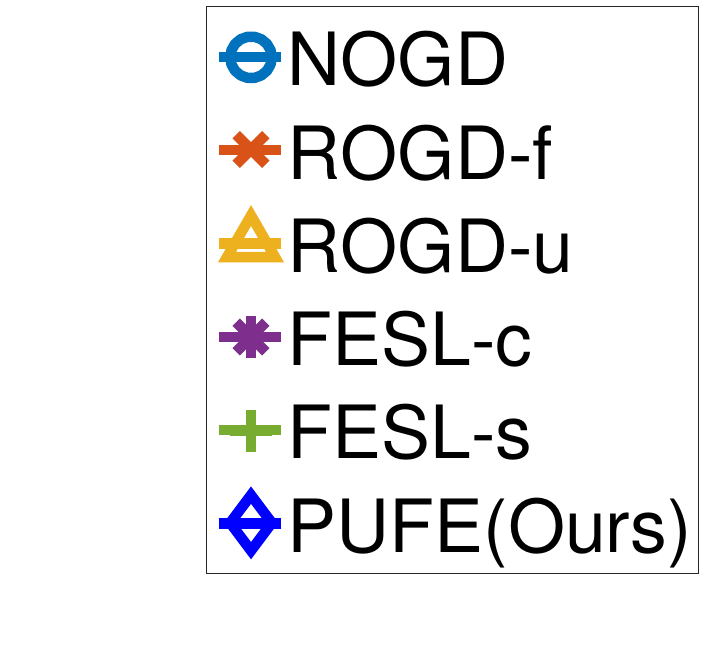}\\
    \vspace{0.1cm}
    \mbox{\scriptsize\quad\quad\quad\emph{legend}}
\end{minipage}
\vspace{-0.5em}
\caption{\small{The trend of average cumulative loss on synthetic and real data. The average cumulative loss is the smaller the better. All the average cumulative loss at any time of our method is comparable to the best baselines. Note that we do not have to be better than the base models and it is sufficient to be comparable with them.}}
\label{fig:synthetic}
\vspace{-1.4em}
\end{figure*}

Figure~\ref{fig:synthetic} gives the trend of average cumulative loss in predictable setting since FESL-c and FESL-s are not comparable in unpredictable setting. (a-i) are the results on synthetic data, (j-k) are the results of the real data. The average cumulative loss is the smaller the better. From the experimental results, we have the following observations. First, NOGD decreases rapidly which conforms to the fact that NOGD on rounds $T_1+1,\ldots,T_1+T_2$ becomes better and better with more and more correct data coming. Besides, ROGD-u also declines but not very apparent since on rounds $1,\ldots,T_1$, ROGD-u already learned well and tend to converge, so updating with more recovered data could not bring too much benefits. Moreover, ROGD-f does not drop down but even go up instead, which is also reasonable because it is fixed and if there are some recovering errors, it will perform worse. FESL-c and FESL-s are based on NOGD and ROGD-u, so their average cumulative losses also decrease. Our PUFE follows the best curve all the time and obtains good performance at the beginning of period $T_1+1,\ldots,T_1+T_2$. This is very important since at the beginning of the current feature space, data are few and a good model is hard to learn but very necessary since for example in ecosystem protection, we need good performance every day or even every single time.

\section{Conclusion}
\label{section:conclusion}
In this paper, we focus on a new and more practical setting: prediction with unpredictable feature evolution. In this setting, we find that the vanishing of old features is usually unpredictable. We attempt to fill this fragmentary period and formulate it as a matrix completion problem. By the free row space obtaining from the preceding matrix, we only need $\Omega(dr\ln r)$ observed entries to recover the target matrix exactly, where $d$ is the row of the target matrix and $r$ is the rank. We also provide a new way to adaptively combine the base models. Theoretical results show that our model is always comparable to the best base model. In this way, at the beginning of the new feature space, our model is still desirable, which conforms to the robustness, an important topic in nowadays machine learning community. 

The data studied in our paper are all tabular data whose features are artificially designed, which limits the application of our method. In the future, we would like to incorporate neural networks that has both feature space transformation and classifier construction~\cite{zhou2021over} to render our method suitable for more sophisticated image and audio tasks. 

\section*{Acknowledgements}
We would like to thank Yuanyu Wan for the valuable discussions, Chiao-Yu Yang for the languange advice, Yu-Hu Yan for the rebuttal and all the reviewers for their constructive feedbacks.



\small
\bibliographystyle{IEEEtran}{
\bibliography{PUFE}}
\vspace{-2cm}
\begin{IEEEbiography}[{\includegraphics[width=1in, height=1.25in, clip,keepaspectratio]{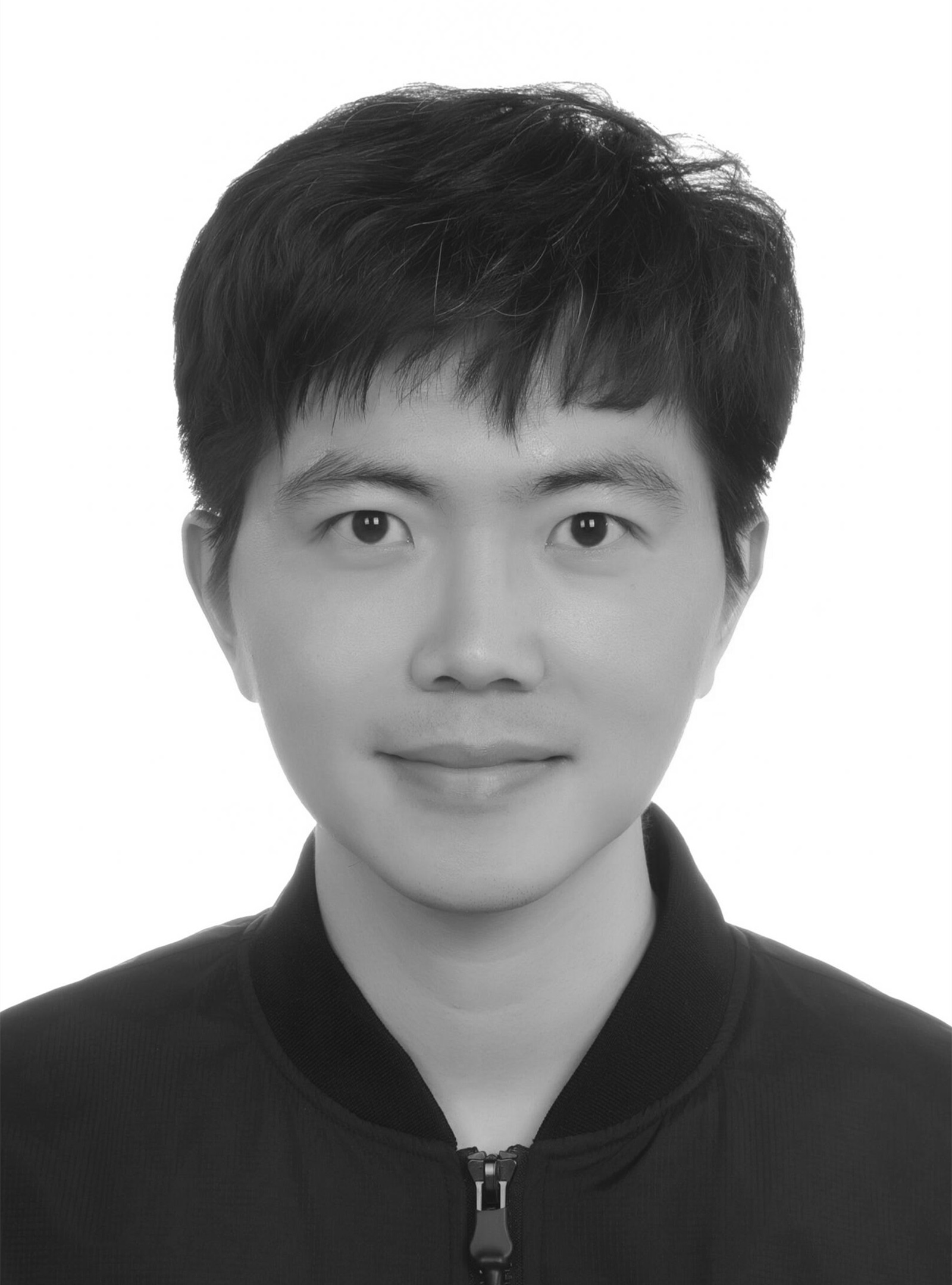}}]{Bo-Jian Hou}
received the B.S. and Ph.D. degrees in the Department of Computer Science \& Technology of Nanjing University, China, in 2014 and 2020, respectively. His main research interests include machine learning and data mining. He won the National Scholarship in 2017. He also won the Program A for Outstanding PhD Candidate of Nanjing University and CCFAI Outstanding Student Paper Award in 2019. In 2020, he won the JSAI Excellent Doctoral Dissertation Award.
\end{IEEEbiography}
\begin{IEEEbiography}[{\includegraphics[width=1in,height=1.25in,clip,keepaspectratio]{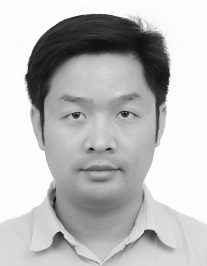}}]{Lijun Zhang}
received the B.S.~and Ph.D.~degrees in Software Engineering and Computer Science from Zhejiang University, China, in 2007 and 2012, respectively. He is currently a Research Professor of the School of Artificial Intelligence, Nanjing University, China. Prior to joining Nanjing University, he was a postdoctoral researcher at the Department of Computer Science and Engineering, Michigan State University, USA. His research interests include machine learning, optimization, information retrieval and data mining. 
\end{IEEEbiography}
\begin{IEEEbiography}[{\includegraphics[width=1in, height=1.25in,clip,keepaspectratio]{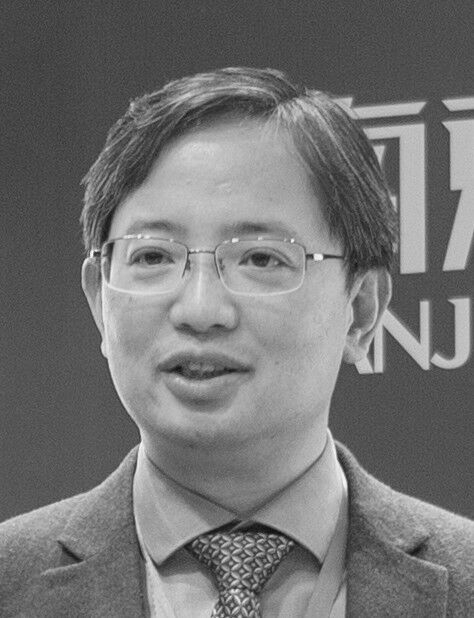}}]{Zhi-Hua Zhou} (S'00-M'01-SM'06-F'13) received the BSc, MSc and PhD degrees in computer science from Nanjing University, China, in 1996, 1998 and 2000, respectively, all with the highest honors. He joined the Department of Computer Science \& Technology at Nanjing University as an Assistant Professor in 2001, and is currently Professor, Head of the Department of Computer Science and Technology, and Dean of the School of Artificial Intelligence; he is also the Founding Director of the LAMDA group. His research interests are mainly in artificial intelligence, machine learning and data mining. He has authored the books \textit{Ensemble Methods: Foundations and Algorithms}, \textit{Evolutionary Learning: Advances in Theories and Algorithms}, \textit{Machine Learning} (in Chinese), and published more than 150 papers in top-tier international journals or conference proceedings. He has received various awards/honors including the National Natural Science Award of China, the IEEE Computer Society Edward J. McCluskey Technical Achievement Award, the CCF-ACM Artificial Intelligence Award, the ACML Distinguished Contribution Award, the PAKDD Distinguished Contribution Award, the IEEE ICDM Outstanding Service Award, the Microsoft Professorship Award, etc. He also holds 24 patents. He is the Editor-in-Chief of the \textit{Frontiers of Computer Science}, Associate Editor-in-Chief of the \textit{Science China Information Sciences}, Action or Associate Editor of the \textit{Machine Learning}, \textit{IEEE Transactions on Pattern Analysis and Machine Intelligence}, \textit{ACM Transactions on Knowledge Discovery from Data}, etc. He served as Associate Editor-in-Chief for \textit{Chinese Science Bulletin} (2008-2014), Associate Editor for \textit{IEEE Transactions on Knowledge and Data Engineering} (2008-2012), \textit{IEEE Transactions on Neural Networks and Learning Systems} (2014-2017), \textit{ACM Transactions on Intelligent Systems and Technology} (2009-2017), Neural Networks (2014-2016), etc. He founded ACML (Asian Conference on Machine Learning), served as Advisory Committee member for IJCAI (2015-2016), Steering Committee member for ICDM, PAKDD and PRICAI, and Chair of various conferences such as General co-chair of ICDM 2016 and PAKDD 2014, Program co-chair of AAAI 2019 and SDM 2013, and Area chair of NeurIPS, ICML, AAAI, IJCAI, KDD, etc. He was the Chair of the IEEE CIS Data Mining Technical Committee (2015-2016), the Chair of the CCF-AI (2012-2019), and the Chair of the CAAI Machine Learning Technical Committee (2006-2015). He is a foreign member of the Academy of Europe, and a Fellow of the ACM, AAAI, AAAS, IEEE, IAPR, IET/IEE, CCF, and CAAI.
\end{IEEEbiography}

\clearpage
\begin{appendix}
    In the supplementary material, we will prove the two theorems in the section ``The Proposed Approach: PUFE''.

\subsection{Proof of Theorem \ref{thm:complete}}
In order to prove that each row $\mathbf{m}_i^\top$ can be recovered exactly, we only need to verify that $V_{\Omega_i}^TV_{\Omega_i}$ is invertible. Therefore, based on the following lemma, we show that our Algorithm 3 with the assumptions in Theorem \ref{thm:complete} satisfies this condition.

\begin{lem}
\label{thm2-4}
For the last row of $M$, with a probability at least $1-\frac{1}{b}e^{-t}$, we have
\begin{eqnarray*}
\lambda_{\min}\left(V_{\Omega_i}^TV_{\Omega_i}\right)\geq\frac{|\Omega_i|}{2d_1}
\end{eqnarray*}
provided that $|\Omega_i|\geq7\mu(r)r(t+\ln r+\ln b)$.
\end{lem}
Then, according to Lemma \ref{thm2-4}
and the union bound, we have
$\lambda_{\min}\left(V_{\Omega_i}^TV_{\Omega_i}\right)\geq\frac{|\Omega_i|}{2d_1}$
with probability at least $ 1-e^{-t}$ for all row $\mathbf{m}_i^\top$ with the fact
$|\Omega_i|=s\geq7\mu(r)r(t+\ln r+\ln b)$. Note that this means all
$V_{\Omega_i}^TV_{\Omega_i}$ are invertible with probability at least $ 1-e^{-t}$. Let $\delta=e^{-t}$, we get $t=\ln(1/\delta)$.

Then we need to prove Lemma~\ref{thm2-4}. Before we do this, we first provide the following supporting lemma.
\begin{lem}
\label{Tropp-2}
(Theorem $2.2$ of \cite{DBLP:journals/aada/Tropp11}) Let $\mathcal{X}$ be a finite set of positive-semidefinite matrices with dimension $k$ (means the size of the square matrix is $k\times k$). $\lambda_{\max}\left(\cdot\right)$ and $\lambda_{\min}\left(\cdot\right)$ calculate the maximum and minimum eigenvalues respectively. Suppose that
\begin{eqnarray*}
\max\limits_{X\in\mathcal{X}}\lambda_{\max}\left(X\right)\leq B.
\end{eqnarray*}
Sample $\{X_1,\cdots,X_\ell\}$ uniformly at random from $\mathcal{X}$ {independently} without replacement. Compute
\begin{eqnarray*}
\mu_{\max}=\ell\cdot\lambda_{\max}\left(\E X_1\right) \quad and\quad 
\mu_{\min}=\ell\cdot\lambda_{\min}\left(\E X_1\right).
\end{eqnarray*}
Then
\begin{align*}
&\Pr\left\{\lambda_{\max}\left(\sum\limits_{i=1}^\ell X_i\right)\geq\left(1+\rho\right)\mu_{\max}\right\}\\
&\leq k\exp\frac{-\mu_{\max}}{B}\left[(1+\rho)\ln(1+\rho)-\rho\right] ~\text{for}~ \rho \geq 0,\\
&\Pr\left\{\lambda_{\min}\left(\sum\limits_{i=1}^\ell X_i\right)\leq\left(1-\rho\right)\mu_{\min}\right\}\\
&\leq k\exp\frac{-\mu_{\min}}{B}\left[(1-\rho)\ln(1-\rho)+\rho\right] ~\text{for}~ \rho \in[0,1).
\end{align*}
\end{lem}

We are now ready to prove Lemma~\ref{thm2-4}.

\begin{proof}[Proof of Lemma~\ref{thm2-4}]
According to the previous definition, $V_{(j)}$, $j\in\Omega_i$ is the $j$-th row vector of $V$.
We have
\begin{eqnarray*}
V_{\Omega_i}^TV_{\Omega_i} = \sum\limits_{j\in\Omega_i}V_{(j)}^TV_{(j)}.
\end{eqnarray*}
It is straightforward to show that
\begin{eqnarray*}
\E\left[V_{(j)}^TV_{(j)}\right]=\frac{1}{d_1}I_{r} \text{ and } \E\left[V_{\Omega_i}^TV_{\Omega_i}\right]=\frac{s}{d_1}I_{r}.
\end{eqnarray*}
To bound the minimum eigenvalue of $V_{\Omega_i}^TV_{\Omega_i}$, we need Lemma \ref{Tropp-2},
where we first need to bound the maximum eigenvalue of $V_{(j)}^TV_{(j)}$, which is a
rank-$1$ matrix, whose eigenvalue
\begin{align*}
\max\limits_{j\in [d_1]} \lambda_{\max}\left(V_{(j)}^TV_{(j)}\right)=\max\limits_{j\in[d_1]}\|V_{(j)}\|_2^2\leq\mu\left(r\right)\frac{{r}}{d_1}
\end{align*}
and
\begin{align*}
\mu_{\min}=|\Omega_i|\cdot\lambda_{\min}\left(\E\left[V_{(1)}^TV_{(1)}\right]\right)=\frac{|\Omega_i|}{d_1}.
\end{align*}
Thus, we have
\begin{align*}
&\Pr\left\{\lambda_{\min}\left(V_{\Omega_i}^TV_{\Omega_i}\right)\leq\left(1-\rho\right)\frac{|\Omega_i|}{d_1}\right\}\\
&\leq {r}\exp\frac{-|\Omega_i|/d_1}{r\mu({r})/d_1}[(1-\rho)\ln(1-\rho)+\rho]\\
&={r}\exp\frac{-|\Omega_i|}{{r}\mu(r)}[(1-\rho)\ln(1-\rho)+\rho].
\end{align*}
By setting $\rho=1/2$, we have
\begin{align*}
\Pr\left\{\lambda_{\min}\left(V_{\Omega_i}^TV_{\Omega_i}\right)\leq\frac{|\Omega_i|}{2d_1}\right\}&\leq
{r}\exp\frac{-|\Omega_i|}{7r\mu(r)}\\
&=re^{-\left|\Omega_i\right|/7{r}\mu(r)}
\end{align*}
where with $|\Omega_i|\geq7\mu(r)r(t+\ln r +\ln b)$, we have $re^{-|\Omega_i|/7r\mu(r)}\leq
\frac{1}{b}e^{-t}$, that is
\begin{eqnarray*}
\Pr\left\{\lambda_{\min}\left(V_{\Omega_i}^TV_{\Omega_i}\right)\geq\frac{|\Omega_i|}{2d_1}\right\}\geq
1-\frac{1}{b}e^{-t}.
\end{eqnarray*}
This is valid for the last row of $M$. For other rows of $M$, the observed entries are always greater than or equal to the observed entries of the last row. Thus, for other rows $V_{\Omega_i}^TV_{\Omega_i}$ is always invertible because 
\[
V_{\Omega_i}^TV_{\Omega_i} = \sum\limits_{j\in\Omega_b}V_{(j)}^TV_{(j)}+\sum_{j\in\Omega_i \& j\notin\Omega_b}V_{(j)}^TV_{(j)}\succ 0,
\]
where $\Omega_b$ represents the sample set of the last row of $M$.
\end{proof}

\subsection{Proof of Theorem~\ref{thm:prediction}}
Theorem~\ref{thm:prediction} is a special case of Theorem 3 from AdaNormalHedge~\cite{DBLP:conf/colt/LuoS15}. Before we prove Theorem~\ref{thm:prediction}, we first give some lemmas that we need.
\begin{lem}
\label{potential}
For any $R\in\mathbb{R},S\geq 0$ and $a\in[-1,1]$, we have
\[
\Phi(R+a,S+|a|)\leq\Phi(R,S)+w(R,S)a+\frac{3|a|}{2(S+1)}.
\]
\end{lem}
\begin{proof}
We first argue that $\Phi(R+a,S+|a|)$, as a function of $a$, is piecewise-convex on $[-1,0]$ and $[0,1]$. Since the value of the function is $1$ when $R+a<0$ and is at least $1$ otherwise. It suffices to only consider the case when $R+a\geq 0$. On the interval $[0,1]$, we can rewrite the exponent (ignoring the constant $\frac{1}{3}$) as:
\[
\frac{(R+a)^2}{S+a}=(S+a)+\frac{(R-S)^2}{S+a}+2(R-S),
\]
which is convex in $a$. Combining with the fact that ``if $g(x)$ is convex then $\exp(g(x))$ is also convex'' proves that $\Phi(R+a,S+|a|)$ is convex on $[0,1]$. Similarly when $a\in[-1,0],$ rewriting the exponent as
\[
\frac{(R+a)^2}{S-a}=(S-a)+\frac{(R+S)^2}{S-a}-2(R+S)
\] 
completes the argument.

Now define function $f(a)=\Phi(R+a,S+|a|)-w(R,S)a$. Since $f(r)$ is clearly also piecewise-convex on $[-1,0]$ and $[0,1]$, we know that the curve of $f(a)$ is below the segment connecting points $(-1,f(-1))$ and $(0,f(0))$ on $[-1,0]$, and also below the segment connecting points $(1,f(1))$ and $(0,f(0))$ on $[0,1]$. This can be mathematically expressed as:
\begin{align*}
&f(a)\\
&\leq\max\{f(0)+(f(0)-f(-1))a,f(0)+(f(1)-f(0))a\}\\
&=f(0)+(f(1)-f(0))|a|,
\end{align*}
where we use the fact $f(-1)=f(1).$ Now by Lemma 2 of NormalHedge.DT~\cite{DBLP:conf/nips/LuoS14}, we have
\begin{align*}
&f(1)-f(0)\\
&=\frac{1}{2}(\Phi(R+1,S+1)+\Phi(R-1,S+1))-\Phi(R,S)\\
&\leq\frac{1}{2}\left(\exp\left(\frac{4}{3(S+1)}\right)-1\right),
\end{align*}
which is at most $\frac{e^{\frac{4}{3}}-1}{2(S+1)}$ since $S$ is nonnegative and $e^x-1\leq\frac{e^c-1}{c}x$ for any $x\in[0,c].$ Noting that $e^\frac{4}{3}-1\leq 3$ completes the proof.
\end{proof}

Though in PUFE, we do not assign prior distribution over base models, which means a uniform prior assigned to them, to be general, we here assume there is a prior distribution $\bm{q}\in\Delta_N$ over $N$ base models. Besides, we do not consider the ``asleep'' situation here, which means $\mathcal{I}_{i,t}=1$ for all $i$ and $t$. Thus, for the convenience of proving, at each round, we set the weight of each base model $\alpha_{i,t}$ to be proportional to $w(R_{i,t-1},S_{i,t-1})$:
\[
\alpha_{i,t}\propto q_i w(R_{i,t-1},S_{i,t-1}),
\] 
instead of
\[
\alpha_{i,t}\propto \I_{i,t}w(R_{i,t-1},S_{i,t-1}).
\] 
We redefine 
\[R_{i,T_2}=\sum_{k=T_1+1}^{T_1+T_2} r_{i,k}, \quad S_{i,T_2}=\sum_{k=T_1+1}^{T_1+T_2} |r_{i,k}|,\]
which are slightly different with the original definitions in the main text. Thus we have the Lemma~\ref{bound of potential} that makes use of Lemma~\ref{potential} to show that the weighted sum of potentials does not increase much and thus the final potential is relatively small.
\begin{lem}
\label{bound of potential}
PUFE ensures $\sum_{i=1}^N q_i\Phi(R_{i,T_2},S_{i,T_2})\leq B=1+\frac{3}{2}\sum_{i=1}^N q_i(1+\ln(1+S_{i,T_2})).$
\end{lem}

\begin{proof}
First note that since PUFE predicts $\alpha_{i,t}\propto q_i w(R_{i,t-1},S_{i,t-1}),$ we have
\begin{equation}
\label{sum of potential}
\sum_{i=1}^N q_i w(R_{i,t-1},S_{i,t-1})r_{i,t}=0.
\end{equation}
Now applying Lemma~\ref{potential} with $R=R_{i,t-1},S=S_{i,t-1}$ and $a=r_{i,t},$ multiplying the inequality by $q_i$ on both sides and summing over $i$ gives $\sum_{i=1}^N q_i\Phi(R_{i,t},S_{i,t})\leq\sum_{i=1}^N q_i\Phi(R_{i,t-1},S_{i,t-1})+\frac{3}{2}\sum_{i=1}^N\frac{q_i|r_{i,t}|}{S_{i,t-1}+1}.$ We then sum over $t\in\{T_1+1,\ldots,T_1+T_2\}$ and telescope to show $\sum_{i=1}^N q_i\Phi(R_{i,T_2},S_{i,T_2})\leq 1+\frac{3}{2}\sum_{i=1}^N q_i\sum_{t=T_1+1}^{T_1+T_2}\frac{|r_{i,t}|}{S_{i,t-1}+1}.$ Finally applying Lemma 14 of ML-Prod~\cite{DBLP:conf/colt/GaillardSE14} to show $\sum_{t=T_1+1}^{T_2}\frac{|r_{i,t}|}{S_{i,t-1}+1}\leq 1+\ln(1+S_{i,T_2})$ completes the proof.
\end{proof}

We are now ready to prove Theorem~\ref{thm:prediction}.
\begin{proof}[Proof of Theorem~\ref{thm:prediction}]
Assume $q_1\Phi(R_{1,T_2},S_{1,T_2})\geq\ldots\geq q_N\Phi(R_{N,T_2},S_{N,T_2})$ without loss of generality. Then by Lemma~\ref{bound of potential}, it must be true that $q_i\Phi(R_{i,T_2},S_{i,T_2})\leq\frac{B}{i}$ for all $i$, which, by solving for $R_{i,T_2},$ gives
\[
R_{i,T_2}\leq\sqrt{3S_{i,T_2}\ln\left(\frac{B}{iq_i}\right)}.
\] 
Multiplying both sides by $u_i,$ summing over $N$ and applying the Cauchy-Schwarz inequality, we arrive at 
\begin{align*}
R(\bm{u})&\leq\sum_{i=1}^N\sqrt{3u_i S_{i,T_2}\cdot u_i\ln\left(\frac{B}{i q_i}\right)}\\
&\leq\sqrt{3(\bm{u}\cdot \bm{S}_{T_2})(D(\bm{u}\|\bm{q})+\ln B)},
\end{align*} 
where we define $D(\bm{u}\|\bm{q})=\sum_{i=1}^N u_i\ln\left(\frac{1}{i q_i}\right).$ It remains to show that $D(\bm{u}\|\bm{q})$ and $\text{RE}(\bm{u}\|\bm{q})$ are close. Indeed, we have $D(\bm{u}\|\bm{q})-\text{RE}(\bm{u}\|\bm{q})=\sum_{i=1}^N u_i\ln\left(\frac{1}{i u_i}\right),$ which, by standard analysis, can be shown to reach its maximum when $u_i\propto\frac{1}{i}$ and the maximum value is $\ln\sum_{i}\frac{1}{i}\leq\ln(1+\ln N).$ Thus we have
\begin{equation}
\label{single-bound}
\begin{split}
&\hat{L}_{T_2}\\
&\leq \bm{u}^\top\bm{L}_{T_2}+\sqrt{3(\bm{u}\cdot\bm{S}_{T_2})(\text{RE}(\bm{u}\|\bm{q})+\ln B+\ln(1+\ln N))}\\
&=\bm{u}^\top\bm{L}_{T_2}+\hat{O}(\sqrt{(\bm{u}\cdot\bm{S}_{T_2})\text{RE}(\bm{u}\|\bm{q})}).
\end{split}
\end{equation}
Then in PUFE, because we do not assign prior distribution to base models, so $\bm{q}$ is a uniform one. Thus the worst case of $\text{RE}(\bm{u}\|\bm{q})$ is $\ln N$. When we go back to the exact prediction rule we use, i.e.,
\[
\alpha_{i,t}\propto \I_{i,t}w(R_{i,t-1},S_{i,t-1}),
\]
it suffices to point out that $r_{i,t}$ is still in the interval $[-1,1]$ and Eq. (\ref{sum of potential}) in the proof of Lemma~\ref{bound of potential} still holds by this new prediction rule. Therefore, adding ``asleep'' parameter $\I_{i,t}$ will not change the result. One thing should be noticed that in this case, we have $N_{T_2}$ base models instead of $N$ where $N_{T_2}$ is the total number of the base models created from $T_1+1$ to $T_2$. Thus the worst case of $\text{RE}(\bm{u}\|\bm{q})$ becomes $\ln N_{T_2}$ and $\ln\ln N$ becomes $\ln\ln N_{T_2}$ (we put it in $\hat{O}$), which completes the proof. 
\end{proof}

\end{appendix}

\end{document}